\documentclass{article}

\PassOptionsToPackage{numbers, compress}{natbib}

\usepackage[main, final]{neurips_2026}

\usepackage[utf8]{inputenc} 
\usepackage[T1]{fontenc}    
\usepackage{hyperref}       
\usepackage{url}            
\usepackage{booktabs}       
\usepackage{amsfonts}       
\usepackage{nicefrac}       
\usepackage{microtype}      
\usepackage{xcolor}         
\usepackage{amsmath}
\usepackage{amsthm}
\usepackage{graphicx}
\usepackage{enumitem}
\usepackage{algorithm}
\usepackage{algorithmic}
\usepackage{multirow}

\usepackage{booktabs}
\usepackage{multirow}
\usepackage{caption}
\usepackage{graphicx}  

\theoremstyle{plain}
\newtheorem{theorem}{Theorem}[section]
\newtheorem{proposition}[theorem]{Proposition}
\newtheorem{lemma}[theorem]{Lemma}

\theoremstyle{definition}
\newtheorem{definition}[theorem]{Definition}
\newtheorem{assumption}[theorem]{Assumption}
\theoremstyle{remark}
\newtheorem{remark}[theorem]{Remark}

\title{Align Before Aggregation: Basis-Consistent Federated LoRA under Heterogeneous Ranks}

\author{%
  \textbf{Pengpeng Qiao}\textsuperscript{1}\quad
  \textbf{Yang Cao}\textsuperscript{1}\quad
  \textbf{Lingling Zhang}\textsuperscript{2}\quad
  \textbf{Guo Cheng}\textsuperscript{3}\\
  \textbf{Junwei Chen}\textsuperscript{1}\quad
  \textbf{Manjiang Yu}\textsuperscript{4}\quad
  \textbf{Wei Yang Bryan Lim}\textsuperscript{5}\quad
  \textbf{Masatoshi Yoshikawa}\textsuperscript{6}\\[0.5em]
  \textsuperscript{1}Institute of Science Tokyo\quad
  \textsuperscript{2}Capital Normal University\\
  \textsuperscript{3}Beijing Institute of Technology\quad
  \textsuperscript{4}The University of Queensland\\
  \textsuperscript{5}Nanyang Technological University\quad
  \textsuperscript{6}Osaka Seikei University\\[0.3em]
  \texttt{peng2qiao@gmail.com, cao@c.titech.ac.jp, 7089@cnu.edu.cn}\\
  \texttt{guo\_cheng@bit.edu.cn, 214wadechen@gmail.com, manjiang.yu@uq.edu.cn}\\
  \texttt{bryan.limwy@ntu.edu.sg, yoshikawa-mas@g.osaka-seikei.ac.jp}\\
}

\begin{document}

\maketitle

\begin{abstract}
Federated fine-tuning of large language models (LLMs) under heterogeneous client capacities is increasingly important, and low-rank adaptation (LoRA) with heterogeneous ranks makes it practical. Existing server-side aggregators compose local matrix updates $\Delta W_i$ from client-specific LoRA factors $(A_i,B_i)$ before aggregating them. Although these matrix updates lie in the same weight space $\mathbb{R}^{m\times n}$ and can be averaged directly, the LoRA factors that generate them may expressed client-specific local basis. In the typical compose--aggregate--factorize--dispatch loop, this basis inconsistency can make the rank-$r_i$ reference prefix dispatched to lower-rank clients unstable after server-side refactorization. In this work, we propose \textbf{FedAbA}, a basis-consistent federated LoRA aggregation framework for heterogeneous ranks. FedAbA first extracts each client's local basis, aligns it to the corresponding prefix of the global reference basis, and constructs a basis-consistent reconstruction of each local matrix update before aggregation. It then aggregates these reconstructions with weights that combine client data size and consistency score, followed by exact coefficient-space SVD refactorization and gauge fixing before rank-prefix dispatch. Our theoretical analysis clarifies the role of basis consistency and supports the proposed consistency-aware aggregation mechanism. Extensive experiments show that FedAbA outperforms representative federated LoRA baselines, with the largest gains under the strongest rank heterogeneity.
\end{abstract}

\section{Introduction}
\label{sec:introduction}
Large pretrained models, especially large language models (LLMs)~\cite{cong2025systematic,chowdhery2023palm,grattafiori2024llama}, have become a central backbone for a wide range of downstream applications~\cite{mohammadi2025evaluation,wang2026metaeval}. In many realistic deployments, however, the data needed to adapt these models is decentralized and privacy-sensitive. Federated learning (FL) offers a natural solution by allowing distributed clients to collaboratively fine-tune a shared model without exposing raw local data~\cite{tang2024z,qiao2024feed,tian2026meta,zhu2026fedrahi,wu2026one,yan2026spattack,bi2024fedsig}. 
Since full-parameter federated fine-tuning is prohibitively expensive in both communication and on-device computation, low-rank adaptation (LoRA)~\cite{hu2022lora} has emerged as a practical parameter-efficient choice for federated fine-tuning of LLMs. In practice, clients also differ substantially in resources~\cite{bai2024federated,fang2026towards,koo2025towards,wang2024flora}, which naturally leads them to adopt different ranks $r_i$, yielding \emph{heterogeneous-rank} federated LoRA.

\begin{figure}[t]
    \centering
    \includegraphics[width=1\textwidth]{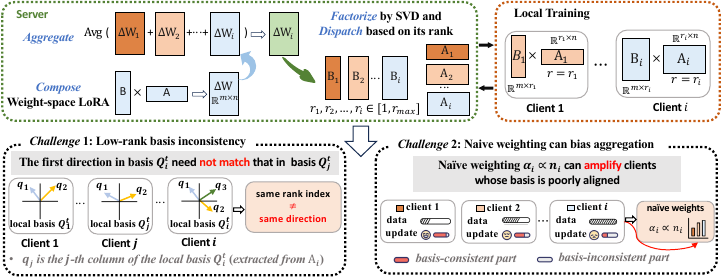}
    \caption{Illustration of federated LoRA under heterogeneous ranks. Compose--aggregate--factorize--dispatch training process (top). Two challenges: basis inconsistency and biased aggregation (bottom).}
    \vspace{-0.2cm}
    \label{fig:intro}
\end{figure}

\textbf{Why heterogeneous ranks are difficult.}
A typical server-side aggregation pipeline in federated LoRA follows a compose--aggregate--factorize--dispatch loop, as shown in Figure~\ref{fig:intro}. Client $i$ uploads LoRA factors $(A_i,B_i)$. The server composes the local matrix update $\Delta W_i=sB_iA_i\in\mathbb{R}^{m\times n}$, averages these local matrix updates across clients, factorizes the averaged matrix into new global LoRA factors, and dispatches the rank-$r_i$ reference prefix of those global factors back to client $i$. The difficulty is not the size of $\Delta W_i$: regardless of $r_i$, every $\Delta W_i$ lies in the same matrix space and can be averaged as a matrix. The difficulty arises from what happens after this matrix average, because the server repeatedly moves between matrix updates and low-rank factors, while the factors uploaded by different clients can be learned in client-specific local basis.
This mismatch comes from the non-uniqueness of low-rank factorization. For any invertible matrix $R\in\mathbb{R}^{r_i\times r_i}$,
\[
B_iA_i=(B_iR)(R^{-1}A_i),
\qquad
A_i\in\mathbb{R}^{r_i\times n},
\qquad
B_i\in\mathbb{R}^{m\times r_i}.
\]
Thus, even two clients that learn similar matrix updates can encode them in different local basis. This ambiguity is less harmful when all clients use the same rank, because each client receives the full global LoRA factors and no rank-prefix truncation is imposed. Under heterogeneous ranks, however, lower-rank clients receive only the rank-$r_i$ reference prefix of the global LoRA factors. After local training, the first uploaded direction of client $i$ need not correspond to the same reference direction as the first uploaded direction of client $j$. When the server aggregates local matrix updates, refactorizes via SVD, and redistributes rank-$r_i$ reference prefixes, the prefix coordinates received by lower-rank clients may change their semantics across rounds. This makes it important to define aggregation relative to an explicit global reference basis. We refer to this phenomenon as \emph{basis inconsistency}.

Furthermore, basis inconsistency also affects how clients should be weighted. The typical aggregation strategy uses data-size weighting. Under heterogeneous-rank LoRA, however, a client with many samples can still be poorly aligned with the global reference basis. Giving such a client a large weight may amplify components that are poorly compatible with the reference basis and are later affected by reconstruction or refactorization. This suggests that server aggregation should reflect not only how much data a client has, but also how consistent its local basis is with the global reference basis.

\textbf{Motivation: Align before aggregation.}
To address these challenges, we propose FedAbA, a \emph{basis-consistent} federated LoRA aggregation framework that aligns client updates to a shared global reference basis before aggregation. Using the orthonormal basis induced by the previous-round global right factor as the reference basis, the server extracts each client's local basis from its uploaded right factor, aligns it to the corresponding reference prefix via orthogonal Procrustes, and reconstructs the local matrix update on that prefix. This reconstruction is essential: merely rotating local LoRA factors leaves the composed update $\Delta W_i$ unchanged and therefore cannot alter server aggregation.
The alignment step also produces a \emph{consistency score} that measures how well the client's local basis matches the reference prefix. FedAbA then aggregates basis-consistent reconstructions using weights that combine client data size and consistency score. Finally, the averaged basis-consistent global update is refactorized by an exact coefficient-space SVD, avoiding a full $m\times n$ SVD, and a lightweight gauge-fixing step removes arbitrary sign flips and near-degenerate ordering changes before the first $r_i$ components are sent back to client $i$.
Our contributions are summarized as follows:

\begin{itemize}[leftmargin=2em]
    \item We identify that the main challenge in heterogeneous-rank federated LoRA is the lack of explicit basis reconciliation against a global reference basis in the compose--aggregate--factorize--dispatch loop, rather than a dimensional mismatch in the local matrix updates. We also show that factor-space rotations alone cannot resolve this issue.
    
    \item We propose \textbf{FedAbA}, a server-side aggregation framework that reconstructs each local matrix update under a shared global reference basis before aggregation, uses both data size and consistency score in aggregation, and enables an exact coefficient-space SVD refresh, rather than a full-update SVD, with gauge fixing before rank-prefix dispatch.
    
    \item We provide theoretical analysis showing how the consistency score characterizes basis compatibility and supports consistency-aware aggregation under heterogeneous ranks. Experiments show that FedAbA consistently outperforms representative federated LoRA baselines.
\end{itemize}

\begin{figure}[t]
    \centering
    \includegraphics[width=0.98\textwidth]{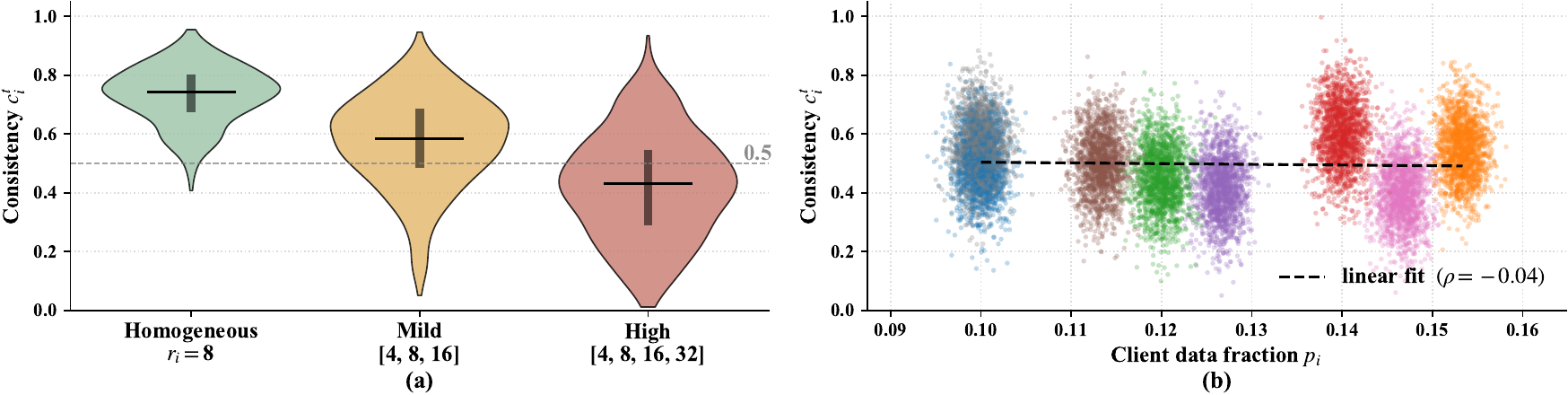}
    \caption{%
(a) Distribution of $c_i^t$ at $t{=}10$ shifts downward as rank heterogeneity grows.
(b) Under strong heterogeneity, $c_i^t$ is weakly correlated with the data-size weight $p_i$.%
}
    \label{fig:motivation}
    \vspace{-0.3cm}
\end{figure}

\section{Method}
\label{sec:method}

\subsection{Problem Setup}
\label{subsec:preliminaries}

We consider federated fine-tuning of LLMs under heterogeneous client LoRA ranks. A server coordinates $N$ clients. Client $i$ owns a private dataset $\mathcal{D}_i$ of size $n_i$ and uses LoRA rank $r_i\in[1,r_{\max}]$. At round $t$, an active subset $\mathcal{S}^t\subseteq\{1,\ldots,N\}$ of clients participates. We describe the method for a single LoRA-inserted layer and omit the layer index.

Let $W_0\in\mathbb{R}^{m\times n}$ be the frozen pretrained weight. Client $i$ maintains local LoRA factors $A_i^t\in\mathbb{R}^{r_i\times n}$ and $B_i^t\in\mathbb{R}^{m\times r_i}$, producing the local matrix update
$
\Delta W_i^t = s\,B_i^t A_i^t \in \mathbb{R}^{m\times n},
$
where $s$ is the LoRA scaling factor. Unless otherwise specified, both $A_i$ and $B_i$ are trainable during local fine-tuning, so the row space of $A_i^t$ and the corresponding local basis may change across rounds. The server maintains global LoRA factors $A_g^t\in\mathbb{R}^{r_{\max}\times n}$ and $B_g^t\in\mathbb{R}^{m\times r_{\max}}$, and distributes the rank-$r_i$ prefix $A_g^t[1\!:\!r_i,:]$ and $B_g^t[:,1\!:\!r_i]$ to client $i$.

Under heterogeneous ranks, lower-rank clients receive only the rank-$r_i$ prefix of the global factors, so the per-coordinate semantics of those prefixes are determined by how the global basis is ordered across rounds. Aggregating client contributions on a shared, explicit global reference basis is therefore central to controlling cross-round prefix stability. We next quantify how severe this basis inconsistency is in practice (Section~\ref{subsec:motivation}) before describing FedAbA (Section~\ref{subsec:method_main}). Table~\ref{tab:notation} summarizes the notation.

\subsection{Motivation}
\label{subsec:motivation}

The basis-inconsistency phenomenon described in Section~\ref{sec:introduction} is naturally quantified by the alignment between each client's local basis and the global reference basis it should be measured against. For client $i$ at round $t$, let $Q_i^t \in \mathbb{R}^{n\times r_i}$ be the local basis obtained from a thin QR decomposition of the uploaded right factor $(A_i^t)^\top$, and let $Q_{\mathrm{ref},i}^{t-1}\in\mathbb{R}^{n\times r_i}$ denote the reference prefix taken from the global reference basis $Q_g^{t-1}=(A_g^{t-1})^\top$. We define the \emph{consistency score}
\begin{equation}
c_i^t
=
\frac{1}{r_i}\,\mathrm{tr}(\Sigma_i^t)
\;\in\;[0,1],
\qquad
(Q_i^t)^\top Q_{\mathrm{ref},i}^{t-1}
=
U_i^t\Sigma_i^t(V_i^t)^\top,
\label{eq:motivation_ci}
\end{equation}
which is the average principal-angle cosine between the local basis and the reference prefix ($c_i^t=1$ means perfect alignment). This is the same quantity that FedAbA later uses for both basis alignment and aggregation weighting (Section~\ref{subsec:method_main}); we therefore measure it directly on a representative federated LoRA training backbone. Note that $c_i^t<1$ generically because (i) $A_i$ is locally trainable, so the local basis a client uploads has drifted from the reference prefix it received at the start of the round, and (ii) the server's SVD-based refactorization can refresh the global basis between rounds.
To probe how this consistency behaves in practice, we fine-tune \textsc{LLaMA-2-7B} with $N{=}8$ clients on \textsc{Dolly}-15K using FlexLoRA~\cite{bai2024federated} as the training backbone (see Appendix~\ref{app:motivation_setup} for details). 

\textbf{Observation 1 (rank heterogeneity sharply degrades basis consistency).}
Figure~\ref{fig:motivation}(a) shows the distribution of $c_i^t$ over all clients and adapted layers at round $t{=}10$ under the three regimes. Even in the homogeneous case the median sits well below $1$, and as rank heterogeneity grows the entire distribution shifts downward, with a substantial mass of $c_i^t$ values falling below $0.5$ under high heterogeneity. By Theorem~\ref{thm:reconstruction_error}, a smaller $c_i^t$ directly translates into a larger reconstruction residual $\sqrt{2r_i(1-c_i^t)}$, so naive aggregation under heterogeneous ranks operates on increasingly inconsistent local bases. This motivates explicitly aligning each client's local basis to the reference prefix \emph{before} aggregation, rather than relying on post-hoc server-side SVD to reconcile mismatched contributions.

\textbf{Observation 2 ($c_i^t$ varies sharply across clients and is weakly correlated with data size).}
Figure~\ref{fig:motivation}(b) plots $c_i^t$ against the data-size weight $p_i$ under the high-heterogeneity regime. The two quantities are weakly correlated ($|\rho|<0.2$), and clients with the largest $p_i$ are not necessarily those with the largest $c_i^t$. Pure data-size weighting can therefore amplify contributions from clients whose local basis is poorly aligned with the reference prefix, exactly the case in which the reconstruction residual is largest. This motivates a \emph{consistency-aware} aggregation weight that combines $p_i$ with $c_i^t$.

\begin{figure}[t]
    \centering
    \includegraphics[width=1\textwidth]{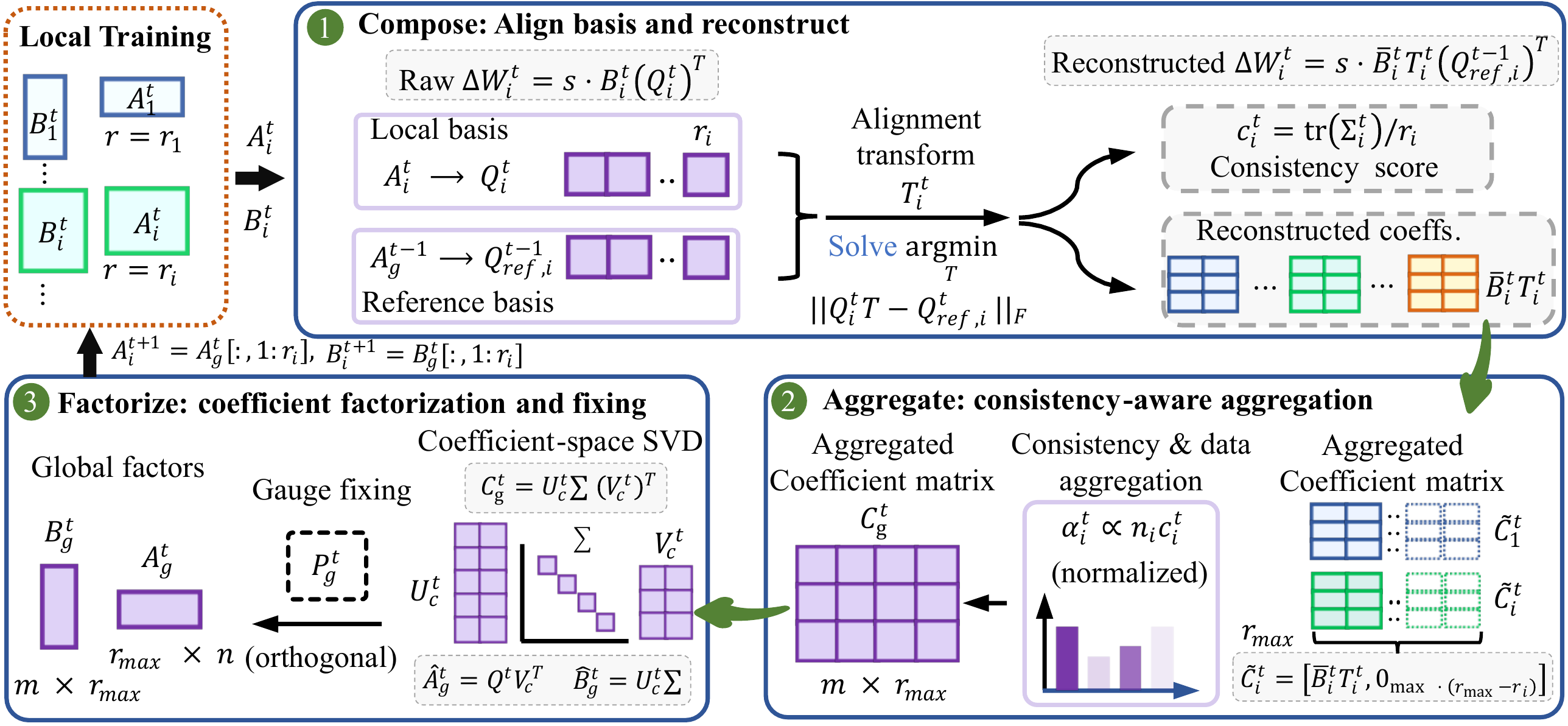}
    \caption{Overview of FedAbA.
}
    \label{fig:framework}
    \vspace{-0.2cm}
\end{figure}

Together, these observations point to two design choices that FedAbA makes in Section~\ref{subsec:method_main}: (i) before aggregation, replace each client's local basis with the corresponding reference prefix via an orthogonal Procrustes alignment, so that all client contributions are expressed under a shared global reference basis; and (ii) weight client contributions by $\alpha_i^t \propto p_i\, c_i^t$, so that data-rich but basis-misaligned clients no longer dominate the aggregate. Fig.~\ref{fig:framework} illustrates the overall workflow of FedAbA.

\subsection{Basis-Consistent Federated LoRA Aggregation}
\label{subsec:method_main}

At the start of round $t$, the server holds the previous-round global LoRA factors $(A_g^{t-1},B_g^{t-1})$. We define the \emph{global reference basis} as
$
Q_g^{t-1}
:=
(A_g^{t-1})^\top
\in\mathbb{R}^{n\times r_{\max}},
$
which has orthonormal columns by construction. For client $i$, the corresponding \emph{reference prefix} is
$
Q_{\mathrm{ref},i}^{t-1}
:=
Q_g^{t-1}[:,1:r_i]
\in\mathbb{R}^{n\times r_i}.
$
All clients in the same round are therefore aligned to prefixes of the same global reference basis.

FedAbA proceeds in three steps: (i) extract and align each client's local basis; (ii) construct a basis-consistent local matrix update under the reference prefix; and (iii) aggregate these basis-consistent updates with consistency-aware weights and refactorize the aggregate for rank-prefix dispatch.

\subsubsection*{Step 1: Local basis extraction and alignment}

The server does not align the raw right factor $A_i^t$ directly, because $A_i^t$ entangles basis directions with coefficient magnitudes. Instead, it extracts the local basis by a thin QR decomposition of $(A_i^t)^\top$:
\begin{equation}
(A_i^t)^\top
=
Q_i^t(C_i^t)^\top,
\qquad
Q_i^t\in\mathbb{R}^{n\times r_i},
\quad
(Q_i^t)^\top Q_i^t=I_{r_i},
\quad
C_i^t\in\mathbb{R}^{r_i\times r_i}.
\label{eq:thin_qr}
\end{equation}
Substituting $A_i^t=C_i^t(Q_i^t)^\top$ into $\Delta W_i^t=sB_i^tA_i^t$ gives
\begin{equation}
\Delta W_i^t
=
s\,\bar B_i^t(Q_i^t)^\top,
\qquad
\bar B_i^t
:=
B_i^tC_i^t
\in\mathbb{R}^{m\times r_i}.
\label{eq:local_update_qc_form}
\end{equation}
Here $Q_i^t$ is the local basis, while $\bar B_i^t$ contains the corresponding coefficients. This representation separates the basis to be aligned from the coefficients to be preserved.
Given $Q_i^t$ and $Q_{\mathrm{ref},i}^{t-1}$, the server computes the orthogonal Procrustes alignment
\begin{equation}
T_i^t
=
\arg\min_{T\in\mathcal O(r_i)}
\|Q_i^tT-Q_{\mathrm{ref},i}^{t-1}\|_F^2,
\label{eq:procrustes}
\end{equation}
where $\mathcal O(r_i)$ is the orthogonal group. Writing
$
(Q_i^t)^\top Q_{\mathrm{ref},i}^{t-1}
=
U_i^t\Sigma_i^t(V_i^t)^\top,
$
the solution is
$
T_i^t=U_i^t(V_i^t)^\top.
$
This alignment removes the best within-rank rotation between the local basis and the reference prefix while preserving coefficient norms.

The singular values $\sigma_{i,j}^t$ of $(Q_i^t)^\top Q_{\mathrm{ref},i}^{t-1}$ are the cosines of the principal angles between the two bases, and the alignment step recovers the same consistency score introduced in Section~\ref{subsec:motivation}:
$
c_i^t
=
\frac{1}{r_i}\mathrm{tr}(\Sigma_i^t)
=
\frac{1}{r_i}\sum_{j=1}^{r_i}\sigma_{i,j}^t
\in[0,1].
$
A larger $c_i^t$ indicates better consistency between the local basis and the reference prefix; FedAbA reuses $c_i^t$ for aggregation weighting in {\bf Step~3}.

\subsubsection*{Step 2: Basis-consistent reconstruction}

A central design decision of FedAbA is that the server does \emph{not} merely rotate local factors. For any orthogonal $T\in\mathcal O(r_i)$,
$
\bar B_i^tT(Q_i^tT)^\top
=
\bar B_i^t(Q_i^t)^\top.
$
Thus, factor rotation leaves the composed matrix update unchanged. Aggregating rotated factors would therefore produce the same matrix aggregate as aggregating the raw $\Delta W_i^t$. Proposition~\ref{prop:reparam_invariance} formalizes this invariance.

Instead, FedAbA constructs a basis-consistent local matrix update by replacing the local basis with the reference prefix:
\begin{equation}
{\Delta \widetilde W}_i^t
=
s\,\bar B_i^tT_i^t(Q_{\mathrm{ref},i}^{t-1})^\top.
\label{eq:reconstructed_update}
\end{equation}
This construction uses all $r_i$ aligned coefficient columns of client $i$, but expresses them on the reference prefix. It does not select a subset of local basis directions.

Let 
$
E_i^t
:=
Q_i^tT_i^t-Q_{\mathrm{ref},i}^{t-1}.
$
Then the raw local matrix update decomposes as
\begin{equation}
\Delta W_i^t
=
{\Delta \widetilde W}_i^t
+
s\,\bar B_i^tT_i^t(E_i^t)^\top.
\label{eq:update_decomposition}
\end{equation}
The difference
$
\Delta W_i^t-{\Delta \widetilde W}_i^t
$
is the reconstruction residual caused by replacing the local basis with the reference prefix. This residual is not a collection of unused basis directions. It quantifies the difference between the raw local update and its reference-basis reconstruction. In FedAbA, only ${\Delta \widetilde W}_i^t$ is aggregated. Theorem~\ref{thm:reconstruction_error} shows that
$
\|\Delta W_i^t-{\Delta \widetilde W}_i^t\|_F
\le
s\|\bar B_i^t\|_2\sqrt{2r_i(1-c_i^t)}.
$

\subsubsection*{Step 3: Consistency-aware aggregation and refactorization}

The server combines the data-size weight $p_i^t = n_i / \sum_{j\in \mathcal S^t} n_j$ with the consistency score $c_i^t$:
\begin{equation}
\alpha_i^t
=
\frac{p_i^t\,c_i^t}{\sum_{j\in\mathcal S^t}p_j^t\,c_j^t}
=
\frac{n_i\,c_i^t}{\sum_{j\in\mathcal S^t}n_j\,c_j^t}.
\label{eq:consistency_weight}
\end{equation}
We assume \(Z^t=\sum_{j\in\mathcal S^t}p_j^t\,c_j^t\) is positive. The case \(Z^t=0\) occurs only when all active clients have \(c_i^t=0\), so the consistency scores provide no weighting signal; we then fall back to \(\alpha_i^t=p_i^t\). The aggregated basis-consistent update is
$
{\Delta \bar W}_g^t
=
\sum_{i\in\mathcal S^t}\alpha_i^t\,{\Delta \widetilde W}_i^t.
$

Equivalently, define the zero-padded coefficient matrix
$
\widetilde C_i^t
=
\big[
\bar B_i^tT_i^t,\;
0_{m\times(r_{\max}-r_i)}
\big]
\in\mathbb{R}^{m\times r_{\max}}.
$
Then we can rewrite
\begin{equation}
{\Delta \bar W}_g^t
=
s\,C_g^t(Q_g^{t-1})^\top,
\qquad
C_g^t
=
\sum_{i\in\mathcal S^t}\alpha_i^t\widetilde C_i^t.
\label{eq:coefficient_aggregate}
\end{equation}
This coefficient-space view makes the heterogeneous-rank structure explicit: client $i$ contributes coefficient columns associated with the first $r_i$ basis directions of the global reference basis.
To redistribute the aggregate in rank-$r_{\max}$ LoRA form, the server performs an exact coefficient-space SVD refresh. Since $C_g^t\in\mathbb{R}^{m\times r_{\max}}$, it suffices to compute
\begin{equation}
C_g^t
=
U_c^t\Sigma_c^t(V_c^t)^\top,
\label{eq:coeff_svd}
\end{equation}
where $V_c^t\in\mathbb{R}^{r_{\max}\times r_{\max}}$ is orthogonal and zero singular values are padded if necessary. Then
\begin{equation}
{\Delta \bar W}_g^t
=
s\,U_c^t\Sigma_c^t(Q_g^{t-1}V_c^t)^\top.
\label{eq:exact_svd_refresh}
\end{equation}
The provisional global factors are
$
\widehat A_g^t
=
(Q_g^{t-1}V_c^t)^\top,
\widehat B_g^t
=
U_c^t\Sigma_c^t.
$
They exactly represent the aggregated basis-consistent update:
$
s\widehat B_g^t\widehat A_g^t
=
{\Delta \bar W}_g^t.
$
Thus, the server decomposes only the coefficient matrix $C_g^t\in\mathbb{R}^{m\times r_{\max}}$ rather than materializing and decomposing the full $m\times n$ update.

\textbf{Gauge fixing.}
The SVD in Eq.~\eqref{eq:coeff_svd} can introduce arbitrary sign flips and swaps among nearly equal singular values. These changes do not affect the composed matrix update, but they can relabel the factor indices sent to clients. We thus use a lightweight gauge-fixing step before rank-prefix dispatch.

Let $\tau_{\mathrm{svd}}>0$ be a relative-gap threshold. After sorting singular values in descending order, we form contiguous near-degenerate blocks by splitting at adjacent relative gaps larger than $\tau_{\mathrm{svd}}$. Within each block, provisional directions in
$\widehat Q_g^t=(\widehat A_g^t)^\top$
are matched to the corresponding columns of $Q_g^{t-1}$ by maximum absolute cosine similarity, with signs chosen to make the matched cosines nonnegative. This yields a block-wise signed permutation matrix $P_g^t$. The final global factors are
\begin{equation}
A_g^t
=
P_g^t\widehat A_g^t,
\qquad
B_g^t
=
\widehat B_g^t(P_g^t)^\top.
\label{eq:gauge_fixing}
\end{equation}
Since $P_g^t$ is orthogonal,
$
sB_g^tA_g^t
=
s\widehat B_g^t\widehat A_g^t
=
\overline{\Delta W}_g^t.
$
Thus, gauge fixing changes only the factor representation, not the composed global matrix update.

\begin{remark}[Span of the global basis]
\label{rem:rotation_invariance}
The SVD refresh changes the global basis but not its span in the default FedAbA algorithm. After gauge fixing,
\[
Q_g^t
=
(A_g^t)^\top
=
Q_g^{t-1}V_c^t(P_g^t)^\top
=
Q_g^{t-1}O_g^t,
\qquad
O_g^t\in\mathcal O(r_{\max}).
\]
\end{remark}
Therefore,
$
\operatorname{span}(Q_g^t)
=
\operatorname{span}(Q_g^{t-1}).
$
By induction, the span of the global basis remains the warm-start span throughout training. FedAbA should therefore be understood as a basis-consistent aggregation method within a warm-started global basis.

\vspace{-0.1cm}
\subsection{Workflow and Complexity}
\label{subsec:complexity}

Algorithm~\ref{alg:fedaba} summarizes the training workflow of FedAbA. FedAbA requires a non-degenerate initial global reference basis. Since a zero LoRA update does not define meaningful basis directions, we initialize the first global LoRA factors by a warm-start phase: the server performs several standard federated LoRA aggregation rounds without basis-consistent reconstruction, computes a rank-$r_{\max}$ truncated SVD of the resulting aggregated matrix update, and uses the resulting factors as $(A_g^0,B_g^0)$. 

The client-side training protocol and communication are unchanged relative to standard federated LoRA fine-tuning. All additional operations are performed on the server. For each active client, thin QR of $(A_i^t)^\top$ costs $O(nr_i^2)$. Forming $(Q_i^t)^\top Q_{\mathrm{ref},i}^{t-1}$ and solving the Procrustes problem costs $O(nr_i^2+r_i^3)$. Computing $\bar B_i^t=B_i^tC_i^t$ and $\bar B_i^tT_i^t$ costs $O(mr_i^2)$. Coefficient aggregation costs $O(|\mathcal S^t|mr_{\max})$. The final SVD is performed on $C_g^t\in\mathbb{R}^{m\times r_{\max}}$, costing $O(mr_{\max}^2+r_{\max}^3)$, rather than on a full $m\times n$ matrix. Communication complexity remains $O(r_i(m+n))$ per client.

\vspace{-0.3cm}
\section{Theoretical Analysis}
\label{sec:theory}

This section analyzes the aggregation of FedAbA. We first show why factor rotation alone cannot resolve basis inconsistency (Section~\ref{subsec:theory_necessity}), then quantify the reconstruction through the consistency score and justify consistency-aware weighting (Section~\ref{subsec:theory_reconstruction}), and finally give an idealized convergence interpretation (Section~\ref{subsec:theory_optimization}). Throughout, we use the basis--coefficient form $\Delta W_i^t=s\bar B_i^t(Q_i^t)^\top$ and the basis-consistent reconstruction ${\Delta \widetilde W}_i^t=s\bar B_i^tT_i^t(Q_{\mathrm{ref},i}^{t-1})^\top$. Full proofs are in Appendix~\ref{app:theory_full}.

\subsection{Basis Alignment Requires Reconstruction, Not Rotation}
\label{subsec:theory_necessity}

A natural first attempt is to rotate each client's LoRA factors by the Procrustes solution before aggregation. This operation is invisible after composing the matrix update.

\begin{proposition}[Factor rotation invariance]
\label{prop:reparam_invariance}
For $T\in\mathcal O(r_i)$, $s\bar B_i^tT(Q_i^tT)^\top=s\bar B_i^t(Q_i^t)^\top=\Delta W_i^t$.
\end{proposition}

FedAbA therefore replaces the aligned local basis with the reference prefix, as in Eq.~\eqref{eq:reconstructed_update} (proof in Appendix~\ref{app:proof_reparam}). The reconstructed updates then admit a single shared-basis representation.

\begin{proposition}[Shared-basis representation]
\label{prop:shared_coord}
Let $\widetilde C_i^t=[\bar B_i^tT_i^t,\,0_{m\times(r_{\max}-r_i)}]$. Then ${\Delta \widetilde W}_i^t=s\widetilde C_i^t(Q_g^{t-1})^\top$ and ${\Delta \bar W}_g^t=s\big(\sum_{i\in\mathcal S^t}\alpha_i^t\widetilde C_i^t\big)(Q_g^{t-1})^\top$, so $\operatorname{rank}({\Delta \bar W}_g^t)\le r_{\max}$.
\end{proposition}
Consequently (proof in Appendix~\ref{app:proof_shared_coord}), the coefficient-space SVD refresh in Eq.~\eqref{eq:exact_svd_refresh} \emph{exactly} refactorizes ${\Delta \bar W}_g^t$, and the gauge fixing in Eq.~\eqref{eq:gauge_fixing} only re-labels factor coordinates. Both steps are bias-free; their formal statements (exact refresh, gauge invariance, and span invariance within the warm-started global basis) are deferred to Appendix~\ref{app:svd_refresh}.

\subsection{Reconstruction Residual and Consistency-Aware Weighting}
\label{subsec:theory_reconstruction}

Let $(Q_i^t)^\top Q_{\mathrm{ref},i}^{t-1}=U_i^t\Sigma_i^t(V_i^t)^\top$ and recall $c_i^t=\frac{1}{r_i}\operatorname{tr}(\Sigma_i^t)$. The next theorem connects this score to both the basis alignment quality and the reconstruction residual.

\begin{theorem}[Reconstruction residual is controlled by basis consistency]
\label{thm:reconstruction_error}
$\min_{T\in\mathcal O(r_i)}\|Q_i^tT-Q_{\mathrm{ref},i}^{t-1}\|_F^2=2r_i(1-c_i^t)$, and
$
\|\Delta W_i^t-{\Delta \widetilde W}_i^t\|_F\le s\|\bar B_i^t\|_2\sqrt{2r_i(1-c_i^t)}.
$
\end{theorem}

Thus $c_i^t$ is exactly the average principal-angle cosine between local basis and reference prefix (proof in Appendix~\ref{app:proof_reconstruction}). The residual vanishes when $c_i^t=1$ and grows monotonically as alignment degrades. 

\begin{assumption}[Uniform coefficient-radius bound]
\label{assump:kappa}
There exists $\kappa_t>0$ such that $s\|\bar B_i^t\|_2\sqrt{r_i}\le\kappa_t$ for every $i\in\mathcal S^t$.
\end{assumption}

\begin{theorem}[Consistency-aware weighting tightens a reconstruction bound]
\label{thm:weight_tightening}
Let $b_i^t:=\kappa_t\sqrt{2(1-c_i^t)}$. Under Assumption~\ref{assump:kappa}, $\|\Delta W_i^t-{\Delta \widetilde W}_i^t\|_F\le b_i^t$. If $\sum_{j\in\mathcal S^t}p_j^t\,c_j^t>0$, the consistency-aware weights $\alpha_i^t=\frac{p_i^t\,c_i^t}{\sum_{j}p_j^t\,c_j^t}$ satisfy $\sum_i\alpha_i^tb_i^t\le\sum_i p_i^t\,b_i^t$.
\end{theorem}

\begin{remark}[Variational view]
\label{rem:variational}
Theorem~\ref{thm:weight_tightening} (proof in Appendix~\ref{app:proof_weight_tightening}) gives a bounded-error justification. Appendix~\ref{app:variational_weighting} shows that $\alpha_i^t$ is also the unique maximizer of a KL-regularized log-consistency objective relative to $p$, providing a complementary variational characterization.
\end{remark}

\subsection{Idealized Convergence Interpretation}
\label{subsec:theory_optimization}

We finally show how the reconstruction step enters a standard smooth non-convex stochastic descent analysis. Define $G^t:=\sum_{i\in\mathcal S^t}\alpha_i^t\Delta W_i^t$ and $\widetilde G^t:={\Delta \bar W}_g^t$. By Proposition~\ref{prop:shared_coord}, $\operatorname{rank}(\widetilde G^t)\le r_{\max}$, so the SVD refresh and gauge fixing preserve $\widetilde G^t$ exactly and contribute no additional bias beyond the reconstruction step itself.

\begin{assumption}[Idealized smooth stochastic descent model]
\label{assump:conv}
$F$ is $L$-smooth, $\mathbb E[G^t\mid W^t]=\nabla F(W^t)$, and $\mathbb E[\|G^t-\nabla F(W^t)\|_F^2\mid W^t]\le\sigma^2$.
\end{assumption}

\begin{theorem}[Convergence with reconstruction bias]
\label{thm:convergence}
Let $e^t:=\widetilde G^t-G^t$ and consider $W^{t+1}=W^t-\eta\widetilde G^t$ with $\eta\le 1/(4L)$. Under Assumption~\ref{assump:conv},
\begin{equation}
\frac{1}{T}\sum_{t=0}^{T-1}\mathbb E\|\nabla F(W^t)\|_F^2\le\frac{4(F(W^0)-F^\star)}{\eta T}+4L\eta\sigma^2+\frac{2+4L\eta}{T}\sum_{t=0}^{T-1}\mathbb E\|e^t\|_F^2,
\label{eq:convergence}
\end{equation}
and $\|e^t\|_F\le\sum_{i\in\mathcal S^t}\alpha_i^t\,s\|\bar B_i^t\|_2\sqrt{2r_i(1-c_i^t)}$.
\end{theorem}

The bound (proof in Appendix~\ref{app:proof_convergence}) isolates FedAbA's reconstruction effect via a measurable bias term controlled by the consistency scores.

\section{Experiment}

\subsection{Setup}

\textbf{Datasets and metrics.}
We evaluate FedAbA on three benchmarks. For \emph{natural language understanding} (NLU), we use four GLUE tasks~\cite{wang2018glue}: MRPC, CoLA, QNLI, and RTE. For \emph{NLG-style instruction tuning}, we train on Dolly-15K~\cite{conover2023free}, Alpaca~\cite{peng2023instruction}, and WizardLM~\cite{xu2023wizardlm}, and evaluate with MMLU or MT-Bench as specified in Table~\ref{tab:flan_heter}. For \emph{natural instructions} (NI), we use Natural Instructions v2~\cite{wang2022super} and report ROUGE-L/ROUGE-1. For NLU and NLG-style experiments, we construct 10-client federated settings with Dirichlet$(0.5)$ partitioning. For NI, each client corresponds to one task; we sample 100 training clients and use a participation rate of 0.1, i.e., 10 clients per round.

\textbf{Baselines}. We compare FedAbA against six baselines for heterogeneous-rank federated LoRA:  FlexLoRA~\cite{bai2024federated}, HetLoRA~\cite{cho2024heterogeneous}, ECLoRA~\cite{ning2025federated}, LoRA-A$^2$~\cite{koo2025towards}, FLoRA~\cite{wang2024flora} and Fed-PLoRA~\cite{zhangheterogeneous}.

\textbf{Implementation Details}.
For NLU/NLG-style/NI experiments, we use RoBERTa-large/LLaMA-2-7B/LLaMA-1B-dj-refine-150B as the frozen backbone and train for 100/50/100 communication rounds, respectively.
LoRA is inserted into attention projection layers. For NLU and NLG-style, client ranks are assigned as [4, 4, 8, 8, 16, 16, 32, 32, 64, 64]. For NI, client ranks are balanced over 4, 8, 16, and 32.
All methods use AdamW with learning rate $2\times10^{-4}$. 
The batch size is 8 for GLUE and NLG-style experiments, and 4 or 8 for Natural Instructions. 
The maximum sequence length is 128 for GLUE and 512 for the LLaMA-based experiments. 
We uses 5 warm-start rounds.

\begin{table}[t]
    \centering
    \newcommand{\tabincell}[2]{\begin{tabular}{@{}#1@{}}#2\end{tabular}}
    \caption{Performance comparison of six baselines across various models and tasks.}
    \label{tab:flan_heter}
    \resizebox{1\linewidth}{!}{
    \begin{tabular}{lcccc cccc cc}
        \toprule
        \multirow{2}{*}{\textbf{Method}}
            & \multicolumn{4}{c}{\textbf{NLU, RoBERTa-large}}
            & \multicolumn{4}{c}{\textbf{NLG-style, LLaMA-2-7B}}
            & \multicolumn{2}{c}{\textbf{NI, LLaMA-1B}} \\
        \cmidrule(lr){2-5} \cmidrule(lr){6-9} \cmidrule(lr){10-11}
           & \textbf{MRPC} & \textbf{CoLA} & \textbf{QNLI} & \textbf{RTE}
           & \textbf{Wizard} & \textbf{Dolly-15K} & \textbf{Alpaca} & \textbf{Alpaca}
           & \multicolumn{2}{c}{\textbf{Natural Instructions v2}}  \\
           & \textbf{F1} & \textbf{MCC} & \textbf{ACC} & \textbf{ACC} & \textbf{(MMLU)} & \textbf{(MMLU)} & \textbf{(MMLU)} & \textbf{(MT-Bench)} & \textbf{Rouge-L} & \textbf{Rouge-1} \\
        \midrule
        HetLORA & 86.6 & 59.8 & 93.8 & 67.5 & 34.5 & 43.8 & 32.3 & 3.57 & 57.0 & 65.3 \\
        FLoRA                 & 87.2 & 61.8 & 93.5 & 71.8 & 35.4 & 44.9 & 34.6 & 3.80 & 57.1 & 65.3 \\
        LoRA-A$^2$   & 88.9 & 62.5 & 93.8 & 79.5 & 35.6 & 45.4 & 35.6 & 4.03 & 57.4 & 65.4 \\
        FlexLoRA              & 88.7 & 62.4 & 93.8 & 78.3 & 36.2 & 45.9 & 36.2 & 4.13 & 57.6 & 65.5 \\
        ECLoRA                & 89.7 & 62.9 & 93.9 & 82.3 & 36.3 & 46.1 & 36.5 & 4.15 & 57.7 & 65.5 \\
        Fed-PLoRA             & 89.5 & 63.1 & 94.0 & 81.8 & 36.6 & 47.2 & 37.3 & 4.23 & 58.1 & 65.8 \\

        FedAbA & \textbf{90.3} & \textbf{63.6} & \textbf{94.2} & \textbf{83.5} & \textbf{37.1} & \textbf{47.8} & \textbf{37.8} & \textbf{4.32} & \textbf{58.4} & \textbf{66.1} \\
        \bottomrule
    \end{tabular}}
    \vspace{-0.3cm}
\end{table}

\subsection{Performance}
\label{subsec:exp_main}

\noindent\textbf{Exp-1: Main Results.}
Table~\ref{tab:flan_heter} compares FedAbA with heterogeneous-rank LoRA baselines on NLU, NLG-style, and NI benchmarks. FedAbA obtains the best reported results across the three task families. The gains are most visible in settings where rank-prefix dispatch is more sensitive to basis drift, while margins are smaller on tasks where most methods already perform strongly. HetLoRA and FLoRA handle rank mismatch through padding or stacking, and FlexLoRA and ECLoRA aggregate reconstructed updates without explicit basis reconciliation. FedAbA improves this pipeline by making aggregation basis-consistent before refactorization.
We further compare FedAbA and Fed-PLoRA using three independent matched runs on six selected metrics.
Appendix~\ref{app:cr_reproducibility} reports the mean and sample standard deviation across runs; FedAbA achieves a higher mean on all six metrics.

\noindent\textbf{Exp-2: Results on different-rank clients.}
Table~\ref{tab:rank_compare} evaluates the dispatched rank-$r_i$ prefixes on each client's local test split. FedAbA improves all rank groups, with the largest gains at $r=4$ and $r=8$. This is consistent with our motivation: low-rank clients receive only short global prefixes and are most sensitive to cross-round basis drift. The smaller gain at higher ranks is also expected because high-rank clients receive a larger portion of the global adapter. These results show that FedAbA improves not only global average performance but also the stability and utility of low-rank prefixes.
We also compare FedAbA and Fed-PLoRA on a fixed subset containing the 20\% of Natural Instructions clients with the lowest average consistency in the default FedAbA run. FedAbA achieves higher average ROUGE-L and ROUGE-1 on this subset (Appendix~\ref{app:cr_low_consistency}).

\begin{table}[t]
    \centering
    \caption{Performance comparison across different LoRA ranks on Dolly and Natural Instructions v2.}
    \label{tab:rank_compare}
    \resizebox{1\linewidth}{!}{
    \begin{tabular}{lcccccc ccccc}
        \toprule
        \multirow{2}{*}{\textbf{Method}}
            & \multicolumn{6}{c}{\textbf{NLG-style, LLaMA-2-7B, Dolly-15K}}
            & \multicolumn{5}{c}{\textbf{NI, LLaMA-1B, Natural Instructions v2}} \\
        \cmidrule(lr){2-7} \cmidrule(lr){8-12}
            & \textbf{Overall (Rouge-L)} & $r=4$ & $r=8$ & $r=16$ & $r=32$ & $r=64$
            & \textbf{Overall} & $r=4$ & $r=8$ & $r=16$ & $r=32$ \\
        \midrule
        HetLORA       & 29.2 & 27.6 & 28.5 & 29.4 & 30.2 & 31.0 & 57.0 & 54.8 & 56.5 & 57.8 & 59.1 \\
        FLoRA         & 28.7 & 26.8 & 27.9 & 28.8 & 30.1 & 30.9 & 57.1 & 54.2 & 56.1 & 58.0 & 59.5 \\
        LoRA-A$^2$    & 30.0 & 28.5 & 29.4 & 30.1 & 30.8 & 31.4 & 57.4 & 55.0 & 56.7 & 58.2 & 59.8 \\
        FlexLoRA      & 30.2 & 28.0 & 29.0 & 30.4 & 31.3 & 32.0 & 57.6 & 55.1 & 56.9 & 58.4 & 60.0 \\
        ECLoRA        & 30.2 & 28.2 & 29.1 & 30.4 & 31.4 & 31.9 & 57.7 & 55.5 & 57.2 & 58.6 & 60.2 \\
        Fed-PLoRA     & 31.3 & 29.7 & 30.8 & 31.5 & 32.1 & 32.6 & 58.1 & 56.6 & 57.9 & 58.8 & 60.5 \\
        \textbf{FedAbA} & \textbf{32.4} & \textbf{31.2} & \textbf{32.0} & \textbf{32.5} & \textbf{33.0} & \textbf{33.4} & \textbf{58.4} & \textbf{57.4} & \textbf{58.6} & \textbf{59.0} & \textbf{60.6} \\
        \bottomrule
    \end{tabular}}
\end{table}

\begin{table}[t]
    \centering
    \caption{Performance comparison across different rank heterogeneity regimes on NLU, NLG, and NI.}
    \label{tab:rank_regime}
    \resizebox{1\linewidth}{!}{
    \begin{tabular}{llc ccccccc}
        \toprule
        \textbf{Task Family} & \textbf{Rank Regime} & \textbf{Rank Setting}
            & \textbf{HetLoRA} & \textbf{FLoRA} & \textbf{LoRA-A$^2$} & \textbf{FlexLoRA}
            & \textbf{ECLoRA} & \textbf{Fed-PLoRA} & \textbf{FedAbA} \\
        \midrule
        \multirow{3}{*}{\textbf{NLU Avg.}}
            & Homo. & $[8] \times 10$              & 76.33 & 77.98 & 80.00 & 79.80 & 80.80 & 80.75 & \textbf{80.97} \\
            & Mild  & $\{4, 8, 16\}$               & 76.65 & 78.43 & 80.73 & 80.17 & 81.35 & 81.28 & \textbf{81.72} \\
            & High  & $\{4, 8, 16, 32, 64\}$       & 76.93 & 78.58 & 81.18 & 80.80 & 82.20 & 82.10 & \textbf{82.90} \\
        \midrule
        \multirow{3}{*}{\textbf{NLG Avg. (MMLU)}}
            & Homo. & $[8] \times 10$              & 36.37 & 37.70 & 38.07 & 38.77 & 38.86 & 39.14 & \textbf{39.30} \\
            & Mild  & $\{4, 8, 16\}$               & 36.60 & 38.17 & 38.57 & 39.00 & 39.16 & 39.51 & \textbf{39.83} \\
            & High  & $\{4, 8, 16, 32, 64\}$       & 36.87 & 38.30 & 38.87 & 39.43 & 39.63 & 40.37 & \textbf{40.90} \\
        \midrule
        \multirow{3}{*}{\textbf{NI Avg. (ROUGE-L/1)}}
            & Homo. & $r=8$                        & 61.05 & 61.10 & 61.00 & 61.40 & 61.40 & 61.55 & \textbf{61.60} \\
            & Mild  & $\{4, 8, 16\}$               & 61.15 & 61.25 & 61.20 & 61.50 & 61.45 & 61.70 & \textbf{61.90} \\
            & High  & $\{4, 8, 16, 32\}$           & 61.15 & 61.20 & 61.40 & 61.55 & 61.60 & 61.95 & \textbf{62.25} \\
        \bottomrule
    \end{tabular}}
\end{table}

\begin{figure}[t]
    \centering
    \includegraphics[width=1\textwidth]{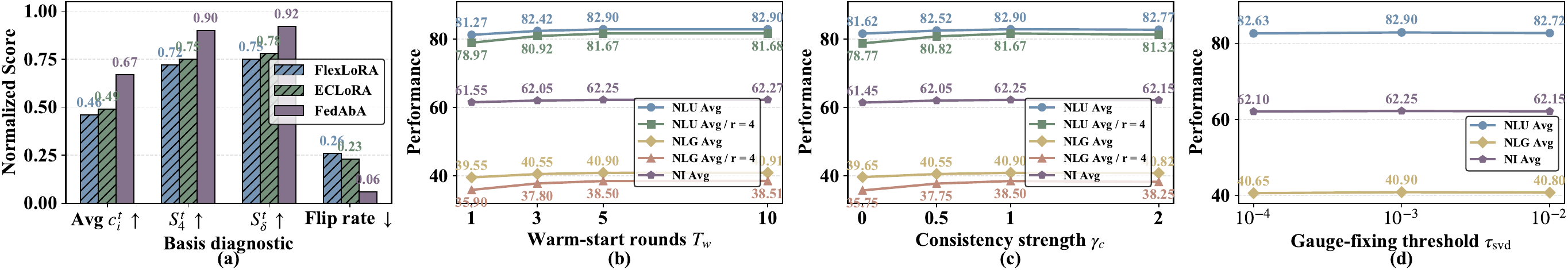}
    \caption{%
Sensitivity analysis under high rank heterogeneity.
(a) Basis diagnostics comparing FlexLoRA, ECLoRA.
(b)--(d) Sensitivity of FedAbA to warm-start rounds $T_w$, consistency weighting strength $\gamma_c$, and gauge-fixing threshold $\tau_{\mathrm{svd}}$, respectively.
}
    \label{fig:sensitivity}
\end{figure}

\begin{table}[htp!]
    \centering
    \begin{minipage}[t]{0.5\linewidth}
        \centering
        \captionof{table}{System efficiency comparison across baselines. Lower is better for all metrics}
        \label{tab:system_efficiency}
        \resizebox{\linewidth}{!}{
        \begin{tabular}{lcccc}
            \toprule
            \multirow{2}{*}{\textbf{Method}}
                & \textbf{Comm.} & \textbf{Effective} & \textbf{Aggregation} & \textbf{Server} \\
                & / Round ($N$=10)        & Rank               & Time          & Memory \\
            \midrule
            HetLoRA         & 520 MB      & 64        & 0.15s & 67 MB \\
            FLoRA  & 2.86 GB     & 248       & 0.12s & 260 MB \\
            LoRA-A$^2$      & 931 MB & 64        & 0.35s & 67 MB \\
            FlexLoRA        & 520 MB      & 64        & 130s  & 2.15 GB \\
            ECLoRA          & 520 MB      & 64        & 4.0s  & 4.3 GB \\
            Fed-PLoRA       & 931 MB      & 64  & 0.35s & 67 MB \\
            \textbf{FedAbA} & \textbf{520 MB} & \textbf{64} & 2.0s & 190 MB \\
            \bottomrule
        \end{tabular}}
    \end{minipage}
    \hfill
    \begin{minipage}[t]{0.486\linewidth}
        \centering
        \captionof{table}{Ablation study of FedAbA components across NLU, NLG, and NI tasks.}
        \label{tab:ablation}
        \resizebox{\linewidth}{!}{
        \begin{tabular}{lccccc}
            \toprule
            \multirow{2}{*}{\textbf{Variant}}
                & \textbf{NLU} & \textbf{NLG} & \textbf{Alpaca} & \textbf{NI} & \textbf{NI} \\
                & Avg.         & MMLU         & MT-Bench        & R-L         & R-1 \\
            \midrule
        w/o basis consistency      & 82.02 & 39.95 & 4.15 & 57.8 & 65.6 \\
        rotation-only              & 82.07 & 40.00 & 4.16 & 57.9 & 65.6 \\
        w/o Procrustes             & 82.32 & 40.20 & 4.20 & 58.0 & 65.8 \\
        w/o $c_i^t$ weighting      & 82.42 & 40.40 & 4.25 & 58.1 & 65.9 \\
        w/o data size              & 82.62 & 40.55 & 4.27 & 58.2 & 66.0 \\
        w/o gauge fixing           & 82.52 & 40.50 & 4.26 & 58.2 & 65.9 \\
        \textbf{FedAbA}            & \textbf{82.90} & \textbf{40.90} & \textbf{4.32} & \textbf{58.4} & \textbf{66.1} \\
        \bottomrule
        \end{tabular}}
    \end{minipage}
    \vspace{-0.2cm}
\end{table}

\noindent\textbf{Exp-3: Effect of Rank Heterogeneity.}
Table~\ref{tab:rank_regime} compares all methods under three rank regimes. In the homogeneous regime, rank-prefix truncation is less exposed because clients use the same LoRA rank. As rank heterogeneity grows, the methods separate more clearly. HetLoRA and FLoRA remain limited by padding or stacking, while FlexLoRA and ECLoRA remain competitive but lack explicit reference-basis reconciliation before aggregation. FedAbA achieves the best performance in all regimes, with its largest margin under high heterogeneity. This trend is consistent with Theorems~\ref{thm:reconstruction_error} and~\ref{thm:weight_tightening}: basis-consistent reconstruction and consistency-aware weighting become more useful when local bases are less consistent with the reference prefix.

\textbf{Exp-4: Basis diagnostics.}
Fig.~\ref{fig:sensitivity}(a) evaluates basis stability under high rank heterogeneity. For each evaluated client, layer, and round, we form
$M_i^t=(Q_i^t)^\top Q_{\mathrm{ref},i}^{t-1}$; its singular values are principal-angle cosines between the uploaded local basis and the reference prefix. We report average consistency $c_i^t$, selected prefix-depth scores $S_k^t=\mathbb{E}[\sigma_k(M_i^t)\mid r_i\ge k]$, and the sign/order flip rate across consecutive global bases. FedAbA yields higher alignment scores and a lower flip rate than baselines.

\textbf{Exp-5: Sensitivity Analysis}.
Figure~\ref{fig:sensitivity}(b)--(d) studies sensitivity under high rank heterogeneity. Performance improves with a short warm-start and saturates around $T_w$=5. For consistency weighting, we use the generalized rule $\alpha_i^t\propto p_i^t(c_i^t)^{\gamma_c}$, where $\gamma_c=1$ is the default setting; moderate consistency weighting performs best. FedAbA is also stable across the tested gauge-fixing thresholds $\tau_{\mathrm{svd}}$.

\textbf{Exp-6: System Efficiency}.
Table~\ref{tab:system_efficiency} compares per-round system cost. FedAbA keeps the same rank-prefix communication as HetLoRA, FlexLoRA, and ECLoRA, and avoids the stacked-rank expansion of FLoRA and auxiliary communication of LoRA-A$^2$/Fed-PLoRA. FedAbA is more expensive than purely factor-space baselines, but substantially cheaper than dense reconstruction-based methods (FlexLoRA and ECLoRA). Its efficiency comes from aggregating coefficient matrices and performing SVD on $C_g^t\in\mathbb{R}^{m\times r_{\max}}$, rather than materializing and decomposing a full $m\times n$ update.

\textbf{Exp-7: Ablation Study}
Table~\ref{tab:ablation} ablates each component on the high-heterogeneity setting, and the order of the drops mirrors the design hierarchy of the method. Removing basis-consistent reconstruction entirely (\emph{w/o basis consistency}) causes the largest drop, identifying reconstruction as the central component. Replacing reconstruction with rotation only (\emph{rotation-only}) recovers almost none of the gain, which empirically confirms Proposition~\ref{prop:reparam_invariance} and rules out factor-rotation strategies as a substitute. Replacing Procrustes with generic orthonormalization loses the optimal alignment guaranteed by Theorem~\ref{thm:reconstruction_error}. Removing $c_i^t$ or the data-size term from the weights shows that the bound-tightening property of Theorem~\ref{thm:weight_tightening} requires both factors. Disabling gauge fixing reintroduces the sign-flip pathology visible in Figure~\ref{fig:sensitivity}(a). Each component thus addresses a distinct, theoretically identified source of basis inconsistency.

\noindent\textbf{Exp-8: Additional Backbones.}
We additionally evaluate FedAbA on Qwen3-4B and LLaMA-2-13B using the Dolly-15K protocol. FedAbA improves MMLU scores over Fed-PLoRA by $0.7$ points on both backbones.
Appendix~\ref{app:cr_backbones} provides the full comparison and experimental setting.

\section{Conclusion}

We proposed FedAbA, a basis-consistent aggregation method for heterogeneous-rank federated LoRA. By aligning client updates before aggregation and stabilizing rank-prefix dispatch, 
FedAbA achieves superior performance compared to SOTA heterogeneous Federated LoRA methods.

\begin{ack}
This work is supported by JSPS KAKENHI 26K21227, JST PRESTO JPMJPR23P5, JST CREST JPMJCR21M2, JST NEXUS JPMJNX25C4, NSFC under Grant 62302043, the Interdisciplinary Project of Capital Normal University under Grant 2026JCZD02, and A*STAR under its Japan-Singapore Joint Call: JST-A*STAR 2024 under Grant R2416IR139.
 
\end{ack}

\bibliographystyle{plainnat}
\bibliography{references}

\newpage

\appendix

\section{Algorithms}
The algorithm of proposed FedAbA is summarized in Algorithm~\ref{alg:fedaba}.

\begin{table}[hpt!]
\centering
\small
\caption{Notation used throughout the paper.}
\label{tab:notation}
\begin{tabular}{ll}
\toprule
Symbol & Meaning \\
\midrule
$N$, $\mathcal{S}^t$ & total number of clients; active client set at round $t$ \\
$n_i$, $p_i^t = n_i/\sum_{j\in \mathcal S^t} n_j$ & client $i$'s sample count; data-size weight \\
$r_i$, $r_{\max}$ & client $i$'s LoRA rank; maximum rank used by the server \\
$(A_i^t, B_i^t)$ & client $i$'s {\bf LoRA factors} at round $t$, shapes $(r_i,n)$ and $(m,r_i)$ \\
$\Delta W_i^t = s B_i^t A_i^t$ & client $i$'s {\bf local matrix update} \\
$(A_g^t, B_g^t)$ & {\bf global LoRA factors} at round $t$, shapes $(r_{\max},n)$ and $(m,r_{\max})$ \\
$Q_g^{t-1} = (A_g^{t-1})^\top$ & \textbf{global reference basis} (round $t$ aggregates against this) \\
$Q_{\mathrm{ref},i}^{t-1} = Q_g^{t-1}[:,1{:}r_i]$ & client $i$'s \textbf{reference prefix} \\
$Q_i^t$ & client $i$'s \textbf{local basis} (thin-QR factor of $(A_i^t)^\top$) \\
$\bar B_i^t = B_i^t C_i^t$ & coefficient matrix in the basis--coefficient form $\Delta W_i^t = s\bar B_i^t (Q_i^t)^\top$ \\
$T_i^t$ & orthogonal Procrustes {\bf alignment matrix} \\
$c_i^t \in [0,1]$ & {\bf consistency score} (between local basis and reference prefix) \\
$\alpha_i^t$ & \textbf{consistency-aware aggregation weight}, $\alpha_i^t \propto p_i\,c_i^t$ \\
${\Delta \widetilde W}_i^t$ & \textbf{basis-consistent reconstruction} of $\Delta W_i^t$ \\
${\Delta \bar W}_g^t$ & {\bf aggregated basis-consistent global update} \\
\bottomrule
\end{tabular}
\end{table}

\begin{algorithm}[hpt!]
\caption{FedAbA: Basis-Consistent Federated LoRA Aggregation}
\label{alg:fedaba}
\begin{algorithmic}[1]
\STATE We assume $Z^t=\sum_j p_j\,c_j^t>0$; the degenerate fallback $\alpha_i^t=p_i$ when $Z^t=0$ is omitted for clarity.
\STATE \textbf{Warm-start:} obtain $(A_g^0,B_g^0)$ by several standard federated LoRA aggregation rounds and rank-$r_{\max}$ SVD.
\FOR{round $t=1,2,\ldots,T$}
    \STATE Server sends $A_g^{t-1}[1:r_i,:]$ and $B_g^{t-1}[:,1:r_i]$ to each active client $i\in\mathcal S^t$.
    \STATE Each client locally fine-tunes both LoRA factors and uploads $(A_i^t,B_i^t)$.
    \STATE Set $Q_g^{t-1}=(A_g^{t-1})^\top$.
    \FOR{each $i\in\mathcal S^t$}
        \STATE Compute thin QR $(A_i^t)^\top=Q_i^t(C_i^t)^\top$ and set $\bar B_i^t=B_i^tC_i^t$.
        \STATE Set $Q_{\mathrm{ref},i}^{t-1}=Q_g^{t-1}[:,1:r_i]$.
        \STATE Compute SVD $(Q_i^t)^\top Q_{\mathrm{ref},i}^{t-1}=U_i^t\Sigma_i^t(V_i^t)^\top$; set $T_i^t=U_i^t(V_i^t)^\top$.
        \STATE Compute $c_i^t=\frac{1}{r_i}\mathrm{tr}(\Sigma_i^t)$.
        \STATE Form $\widetilde C_i^t=[\bar B_i^tT_i^t,\,0_{m\times(r_{\max}-r_i)}]$.
    \ENDFOR
    \STATE \textbf{For all} $i\in\mathcal S^t$: $\alpha_i^t=\frac{p_i^t\,c_i^t}{\sum_{j\in\mathcal S^t}p_j^t\,c_j^t}$.
    \STATE Aggregate $C_g^t=\sum_{i\in\mathcal S^t}\alpha_i^t\widetilde C_i^t$.
    \STATE Compute coefficient-space SVD $C_g^t=U_c^t\Sigma_c^t(V_c^t)^\top$; set $\widehat A_g^t=(Q_g^{t-1}V_c^t)^\top$, $\widehat B_g^t=U_c^t\Sigma_c^t$.
    \STATE Compute block-wise signed permutation $P_g^t$ via SVD gauge fixing (Section~\ref{subsec:method_main}); set $A_g^t=P_g^t\widehat A_g^t$, $B_g^t=\widehat B_g^t(P_g^t)^\top$.
\ENDFOR
\end{algorithmic}
\end{algorithm}

\section{Details of the Motivation Experiments}
\label{app:motivation_setup}

\paragraph{Setup.}
Both observations in Section~\ref{subsec:motivation} use the same federated LoRA training backbone. We fine-tune \textsc{LLaMA-2-7B} with LoRA adapters (rank-scaled by $\alpha/r$ with $\alpha{=}16$) inserted into \texttt{q\_proj} and \texttt{v\_proj} of every transformer block, giving $64$ adapted layers. The \textsc{Dolly}-15K dataset is partitioned across $N{=}8$ clients via a Dirichlet split with concentration $0.5$ over instruction categories; all clients participate in every round. Each round runs one local epoch of AdamW (lr $2\!\times\!10^{-4}$, batch size $8$, sequence length $512$) followed by FlexLoRA~\cite{bai2024federated} server aggregation (compose $\Delta W_i = sB_iA_i$, average with data-size weights, rank-$r_{\max}$ truncated SVD, prefix dispatch). We deliberately use a naive aggregator as the training backbone so the measured quantities reflect the behavior of existing methods, not of FedAbA. Three rank regimes are studied: homogeneous $r_i{=}8$; mild $(4,4,8,8,8,8,16,16)$; strong $(4,4,8,8,16,16,32,32)$. All runs use the same random seed so that differences across regimes are attributable to rank configuration alone. Training runs on $4\times$A100 40\,GB with FSDP, taking $\approx 6$ minutes per round.

\paragraph{Measurements.}
Both subplots use the consistency score from Eq.~\eqref{eq:motivation_ci}. At each round $t$ and each adapted layer $\ell$, we save every active client's uploaded right factor $A_{i,\ell}^t$ and the server's previous-round global right factor $A_{g,\ell}^{t-1}$; we then compute
$Q_{i,\ell}^t$ as the thin-QR $Q$-factor of $(A_{i,\ell}^t)^\top$,
$Q_{\mathrm{ref},i,\ell}^{t-1}=(A_{g,\ell}^{t-1})^\top[:,1:r_i]$, and
$c_{i,\ell}^t = \frac{1}{r_i}\sum_k\sigma_k\big((Q_{i,\ell}^t)^\top Q_{\mathrm{ref},i,\ell}^{t-1}\big)$.

For Observation~1 (subplot~(a)), we record $c_{i,\ell}^{10}$ at round $t{=}10$ for every client $i$ and every adapted layer $\ell$, giving $N\times 64 = 512$ values per regime, which are aggregated into the violin in subplot~(a). An intermediate round is chosen because $A_i^t$ is not yet informative before round $\sim 5$ and FlexLoRA has partially converged after round $\sim 30$; the qualitative trend is stable for any $t\in[8,20]$.

For Observation~2 (subplot~(b)), we run the strong-heterogeneity regime for $T{=}50$ rounds and, for every round $t\in[5,30]$ and every layer $\ell$, plot $(p_i, c_{i,\ell}^t)$ as a single point, where $p_i=n_i/\sum_j n_j$ is the (round-independent) data-size weight. The reported correlation $\rho$ is the Pearson correlation between $p_i$ and $c_{i,\ell}^t$ over all such points. All QR and SVD operations are done in double precision on CPU after detaching from the training graph.

\section{Additional Related Works}
\section{Related Work}
\label{sec:related_work}

\subsection{LoRA with Adaptive Rank}
LoRA~\cite{hu2022lora,che2026lora} reduces the trainable parameters of large pretrained models~\cite{li2025multimodal,naveed2025comprehensive} by representing each weight update as a low-rank product. Since a fixed rank may be suboptimal across layers, tasks, or clients, several works study adaptive-rank LoRA~\cite{huang2025sola,kim2024ra,fan2025make,mao2024dora,chen2025sensitivity,zhang2025more}. AdaLoRA~\cite{zhang2023adalora} allocates rank budgets according to parameter importance under a global parameter constraint, while subsequent LoRA variants explore routing, pruning, or decomposing LoRA components to improve parameter efficiency~\cite{liu2024alora,mao2024dora}. These methods mainly focus on centralized fine-tuning or single-model adaptation, where rank allocation is optimized within one training process.

\subsection{Federated Learning with LoRA}
Federated fine-tuning with LoRA~\cite{bian2026fedalt,hu2022lora,zhang2024autolora,yang2025mtl,he2025cl,lu2026conflict,chen2026privacy} has been widely studied to reduce the communication and computation cost of adapting large language models. A direct application of FedAvg to LoRA factors averages the two factors separately, but this does not generally match the average of the composed updates \(B_iA_i\), leading to aggregation discordance~\cite{wang2024flora}. To avoid this issue, FFA-LoRA freezes one LoRA factor and aggregates the other, while FedSA-LoRA~\cite{guo2024selective} analyzes the asymmetric roles of the LoRA factors and shares only the factor associated with general knowledge. LoRA-A$^2$~\cite{koo2025towards} alternates which factor is trained and combines this with adaptive rank selection. FRLoRA~\cite{yan2025federated} instead accumulates residual low-rank updates into the model weights and reinitializes local LoRA factors in the principal singular space to mitigate constrained parameter space and client drift. RFedLR~\cite{fang2026towards} studies robust federated PEFT under noisy labels, combining sensitivity-aware local tuning with adaptive federated LoRA aggregation.

Another line of work targets heterogeneous client ranks~\cite{cho2024heterogeneous,gao2025federated,wu2026adaptive}. HetLoRA~\cite{cho2024heterogeneous} supports different client ranks through zero-padding and truncation. FLoRA~\cite{wang2024flora} stacks client LoRA factors so that the product of the stacked factors exactly represents the sum of local updates, naturally supporting heterogeneous ranks but increasing the effective global rank. FlexLoRA~\cite{bai2024federated} and ECLoRA~\cite{ning2025federated} compose each local update into a full-size matrix, aggregate these matrix updates, and redistribute rank-specific factors by SVD, thereby leveraging heterogeneous client resources. Fed-PLoRA~\cite{zhangheterogeneous} further reformulates LoRA as multiple parallel one-rank adapters to reduce initialization and aggregation noise under heterogeneous client ranks.
These methods address rank mismatch and aggregation discordance, but they do not explicitly reconcile the client-specific local bases against a shared reference before aggregation. FedAbA is complementary: it first reconstructs each local update under a shared global reference basis, weights updates by both data size and basis consistency, and then performs an exact coefficient-space SVD refresh with gauge fixing for stable rank-prefix dispatch.

\section{Theoretical Proofs}
\label{app:theory_full}

This appendix provides complete statements and proofs for the theoretical results in Section~\ref{sec:theory}, together with two formal results deferred from the main text: the exact-SVD-refresh, gauge-invariance, and span-invariance properties (Section~\ref{app:svd_refresh}, referenced from Section~\ref{subsec:theory_necessity}) and the variational characterization of the consistency-aware weighting rule (Section~\ref{app:variational_weighting}, referenced from Remark~\ref{rem:variational}). We restate each proposition, theorem, and assumption before its proof so that the appendix is self-contained.

\paragraph{Notation.}
After the thin QR decomposition in Eq.~\eqref{eq:thin_qr}, the local matrix update of client $i$ at round $t$ can be written as
\begin{equation}
\Delta W_i^t
=
s\,\bar B_i^t(Q_i^t)^\top,
\qquad
(Q_i^t)^\top Q_i^t=I_{r_i},
\label{eq:app_basis_coeff}
\end{equation}
where $\bar B_i^t=B_i^tC_i^t$. The reference prefix for client $i$ is
\begin{equation}
Q_{\mathrm{ref},i}^{t-1}
=
Q_g^{t-1}[:,1:r_i].
\label{eq:app_ref_prefix}
\end{equation}
The orthogonal Procrustes alignment is
\begin{equation}
T_i^t
=
\arg\min_{T\in\mathcal O(r_i)}
\|Q_i^tT-Q_{\mathrm{ref},i}^{t-1}\|_F^2.
\label{eq:app_procrustes}
\end{equation}
The basis-consistent local matrix update is
\begin{equation}
{\Delta \widetilde W}_i^t
=
s\,\bar B_i^tT_i^t(Q_{\mathrm{ref},i}^{t-1})^\top.
\label{eq:app_reconstruction}
\end{equation}
Writing
\begin{equation}
(Q_i^t)^\top Q_{\mathrm{ref},i}^{t-1}
=
U_i^t\Sigma_i^t(V_i^t)^\top,
\label{eq:app_cross_svd}
\end{equation}
the consistency score is
\begin{equation}
c_i^t
=
\frac{1}{r_i}\operatorname{tr}(\Sigma_i^t)
=
\frac{1}{r_i}\sum_{j=1}^{r_i}\sigma_{i,j}^t.
\label{eq:app_consistency_score}
\end{equation}
Unless otherwise stated, we assume that $Q_i^t$ and $Q_{\mathrm{ref},i}^{t-1}$ have orthonormal columns. We also assume
\[
\sum_{j\in\mathcal S^t}p_j^t\,c_j^t>0
\]
whenever the consistency-aware weight
\[
\alpha_i^t
=
\frac{p_i^t\,c_i^t}{\sum_{j\in\mathcal S^t}p_j^t\,c_j^t}
\]
is used.

\subsection{Factor Rotation Invariance (Proposition~\ref{prop:reparam_invariance})}
\label{app:proof_reparam}

\begin{proposition}[Factor rotation invariance; restatement of Proposition~\ref{prop:reparam_invariance}]
\label{app:prop_reparam}
For any orthogonal matrix $T\in\mathcal O(r_i)$,
\begin{equation}
s\,\bar B_i^tT(Q_i^tT)^\top
=
s\,\bar B_i^t(Q_i^t)^\top
=
\Delta W_i^t.
\label{eq:app_reparam_invariance}
\end{equation}
\end{proposition}

\begin{proof}
Since $T\in\mathcal O(r_i)$, we have $TT^\top=I_{r_i}$. Therefore,
\begin{align}
s\,\bar B_i^tT(Q_i^tT)^\top
&=
s\,\bar B_i^tT\,T^\top(Q_i^t)^\top
\\
&=
s\,\bar B_i^t(TT^\top)(Q_i^t)^\top
\\
&=
s\,\bar B_i^t(Q_i^t)^\top
\\
&=
\Delta W_i^t,
\end{align}
where the last equality follows from Eq.~\eqref{eq:app_basis_coeff}.
\end{proof}

Proposition~\ref{app:prop_reparam} shows that rotating both the coefficient matrix and the local basis is only a reparameterization. It changes the LoRA factors but leaves the composed matrix update unchanged. Therefore, merely rotating local factors before aggregation cannot change the server-side aggregate; FedAbA must explicitly replace the local basis with the reference prefix, as in Eq.~\eqref{eq:app_reconstruction}.

\subsection{Shared-Basis Representation (Proposition~\ref{prop:shared_coord})}
\label{app:proof_shared_coord}

\begin{lemma}[Nested prefix consistency]
\label{app:lem_nested_prefix}
For any two active clients $i,j\in\mathcal S^t$ with $r_i\le r_j$,
\begin{equation}
Q_{\mathrm{ref},i}^{t-1}
=
Q_{\mathrm{ref},j}^{t-1}[:,1:r_i].
\label{eq:app_nested_prefix}
\end{equation}
\end{lemma}

\begin{proof}
By definition,
\[
Q_{\mathrm{ref},i}^{t-1}=Q_g^{t-1}[:,1:r_i],
\qquad
Q_{\mathrm{ref},j}^{t-1}=Q_g^{t-1}[:,1:r_j].
\]
Since $r_i\le r_j$, the first $r_i$ columns of $Q_{\mathrm{ref},j}^{t-1}$ are exactly the first $r_i$ columns of $Q_g^{t-1}$, which are $Q_{\mathrm{ref},i}^{t-1}$.
\end{proof}

\begin{proposition}[Shared-basis representation; restatement of Proposition~\ref{prop:shared_coord}]
\label{app:prop_shared_coord}
Define the zero-padded coefficient matrix
\begin{equation}
\widetilde C_i^t
=
\big[
\bar B_i^tT_i^t,\;
0_{m\times(r_{\max}-r_i)}
\big]
\in\mathbb R^{m\times r_{\max}}.
\label{eq:app_padded_coeff}
\end{equation}
Then
\begin{equation}
{\Delta \widetilde W}_i^t
=
s\,\widetilde C_i^t(Q_g^{t-1})^\top,
\qquad
{\Delta \bar W}_g^t
=
s\left(
\sum_{i\in\mathcal S^t}
\alpha_i^t\widetilde C_i^t
\right)
(Q_g^{t-1})^\top.
\label{eq:app_shared_coord}
\end{equation}
Consequently,
\begin{equation}
\operatorname{rank}({\Delta \bar W}_g^t)\le r_{\max}.
\label{eq:app_rank_consequence}
\end{equation}
\end{proposition}

\begin{proof}
Since
\[
Q_{\mathrm{ref},i}^{t-1}
=
Q_g^{t-1}[:,1:r_i],
\]
we can write
\begin{equation}
(Q_{\mathrm{ref},i}^{t-1})^\top
=
\big[
I_{r_i}\;\;0_{r_i\times(r_{\max}-r_i)}
\big]
(Q_g^{t-1})^\top.
\label{eq:app_selector}
\end{equation}
Substituting Eq.~\eqref{eq:app_selector} into Eq.~\eqref{eq:app_reconstruction},
\begin{align}
{\Delta \widetilde W}_i^t
&=
s\,\bar B_i^tT_i^t(Q_{\mathrm{ref},i}^{t-1})^\top
\\
&=
s\,\bar B_i^tT_i^t
\big[
I_{r_i}\;\;0_{r_i\times(r_{\max}-r_i)}
\big]
(Q_g^{t-1})^\top
\\
&=
s\,
\big[
\bar B_i^tT_i^t,\;
0_{m\times(r_{\max}-r_i)}
\big]
(Q_g^{t-1})^\top
\\
&=
s\,\widetilde C_i^t(Q_g^{t-1})^\top.
\end{align}
Aggregating over active clients gives
\begin{align}
{\Delta \bar W}_g^t
&=
\sum_{i\in\mathcal S^t}
\alpha_i^t{\Delta \widetilde W}_i^t
\\
&=
s
\left(
\sum_{i\in\mathcal S^t}
\alpha_i^t\widetilde C_i^t
\right)
(Q_g^{t-1})^\top.
\end{align}
Let
\[
C_g^t
=
\sum_{i\in\mathcal S^t}
\alpha_i^t\widetilde C_i^t
\in\mathbb R^{m\times r_{\max}}.
\]
Then
\[
{\Delta \bar W}_g^t
=
sC_g^t(Q_g^{t-1})^\top,
\]
so $\operatorname{rank}({\Delta \bar W}_g^t)\le r_{\max}$.
\end{proof}

Proposition~\ref{app:prop_shared_coord} is a per-round statement: at round $t$, all reconstructed client updates are expressed in the global reference basis $Q_g^{t-1}$. After aggregation, the server may refactorize ${\Delta \bar W}_g^t$ and obtain a refreshed basis $Q_g^t$, which becomes the global reference basis for the next round.

\subsection{Exact SVD Refresh, Gauge Invariance, and Span Invariance}
\label{app:svd_refresh}

This subsection provides the formal statements deferred from Section~\ref{subsec:theory_necessity}. It establishes that the SVD refresh in Eq.~\eqref{eq:exact_svd_refresh} and the gauge fixing in Eq.~\eqref{eq:gauge_fixing} together preserve the aggregated update exactly and keep the global basis within its warm-started span, justifying the ``bias-free'' claim used in Section~\ref{subsec:theory_optimization}.

\begin{proposition}[Exact SVD refresh]
\label{app:prop_exact_svd_refresh}
Let
\[
C_g^t
=
\sum_{i\in\mathcal S^t}
\alpha_i^t\widetilde C_i^t
\in\mathbb R^{m\times r_{\max}},
\]
and let its SVD be
\begin{equation}
C_g^t
=
U_c^t\Sigma_c^t(V_c^t)^\top.
\label{eq:app_coeff_svd}
\end{equation}
If
\begin{equation}
\widehat A_g^t
=
(Q_g^{t-1}V_c^t)^\top,
\qquad
\widehat B_g^t
=
U_c^t\Sigma_c^t,
\label{eq:app_exact_svd_factors}
\end{equation}
then
\begin{equation}
s\,\widehat B_g^t\widehat A_g^t
=
{\Delta \bar W}_g^t.
\label{eq:app_exact_svd_update}
\end{equation}
\end{proposition}

\begin{proof}
By Proposition~\ref{app:prop_shared_coord},
\[
{\Delta \bar W}_g^t
=
sC_g^t(Q_g^{t-1})^\top.
\]
Substituting the SVD of $C_g^t$,
\begin{align}
{\Delta \bar W}_g^t
&=
sU_c^t\Sigma_c^t(V_c^t)^\top(Q_g^{t-1})^\top
\\
&=
sU_c^t\Sigma_c^t(Q_g^{t-1}V_c^t)^\top
\\
&=
s\,\widehat B_g^t\widehat A_g^t.
\end{align}
The SVD refresh therefore exactly refactorizes the reconstructed aggregate, with no additional projection error.
\end{proof}

\begin{definition}[Block-wise SVD gauge fixing]
\label{app:def_gauge_fixing}
Let $\widehat Q_g^t=(\widehat A_g^t)^\top$ be the provisional right basis after SVD refresh. The SVD is sign-ambiguous and, inside nearly degenerate singular-value blocks, order-ambiguous. Given a relative threshold $\tau_{\mathrm{svd}}>0$, we form blocks of indices whose singular values are nearly equal. Within each block, we match provisional directions in $\widehat Q_g^t$ to directions in $Q_g^{t-1}$ by maximum absolute cosine similarity and flip signs so that matched cosines are nonnegative. This produces a block-wise signed permutation matrix $P_g^t$.
\end{definition}

\begin{proposition}[Gauge-fixing invariance]
\label{app:prop_gauge_invariance}
Let $(\widehat A_g^t,\widehat B_g^t)$ be provisional global factors satisfying $s\,\widehat B_g^t\widehat A_g^t={\Delta \bar W}_g^t$. Let $P_g^t$ be any orthogonal matrix, including the block-wise signed permutation from Definition~\ref{app:def_gauge_fixing}, and define
\begin{equation}
A_g^t
=
P_g^t\widehat A_g^t,
\qquad
B_g^t
=
\widehat B_g^t(P_g^t)^\top.
\label{eq:app_gauge_factors}
\end{equation}
Then
\begin{equation}
sB_g^tA_g^t
=
s\widehat B_g^t\widehat A_g^t
=
{\Delta \bar W}_g^t.
\label{eq:app_gauge_invariance}
\end{equation}
\end{proposition}

\begin{proof}
Since $P_g^t$ is orthogonal, $(P_g^t)^\top P_g^t=I$. Therefore,
\begin{align}
sB_g^tA_g^t
&=
s\widehat B_g^t(P_g^t)^\top P_g^t\widehat A_g^t
\\
&=
s\widehat B_g^t\widehat A_g^t
\\
&=
{\Delta \bar W}_g^t.
\end{align}
\end{proof}

The gauge-fixing step therefore changes only the factor representation, not the composed global matrix update.

\begin{proposition}[Span invariance of the default FedAbA refresh; formal version of Remark~\ref{rem:rotation_invariance}]
\label{app:prop_span_invariance}
Let
\[
\widehat A_g^t=(Q_g^{t-1}V_c^t)^\top
\]
be the provisional right factor after coefficient-space SVD refresh, and let
\[
A_g^t=P_g^t\widehat A_g^t
\]
be the gauge-fixed right factor, where both $V_c^t$ and $P_g^t$ are orthogonal. Then
\[
Q_g^t:=(A_g^t)^\top
=
Q_g^{t-1}O_g^t
\]
for some $O_g^t\in\mathcal O(r_{\max})$. Consequently,
\[
\operatorname{span}(Q_g^t)
=
\operatorname{span}(Q_g^{t-1}).
\]
\end{proposition}

\begin{proof}
By definition,
\[
Q_g^t
=
(A_g^t)^\top
=
(P_g^t\widehat A_g^t)^\top
=
(\widehat A_g^t)^\top(P_g^t)^\top.
\]
Since $(\widehat A_g^t)^\top=Q_g^{t-1}V_c^t$,
\[
Q_g^t
=
Q_g^{t-1}V_c^t(P_g^t)^\top.
\]
Let $O_g^t:=V_c^t(P_g^t)^\top$. Because both $V_c^t$ and $P_g^t$ are orthogonal, $O_g^t$ is also orthogonal. Hence $Q_g^t=Q_g^{t-1}O_g^t$, which implies $\operatorname{span}(Q_g^t)=\operatorname{span}(Q_g^{t-1})$.
\end{proof}

This proposition shows that the default FedAbA refresh changes the global basis only within the span of the previous global basis. It does not introduce new basis directions after warm-start.

\subsection{Reconstruction Residual (Theorem~\ref{thm:reconstruction_error})}
\label{app:proof_reconstruction}

\begin{theorem}[Reconstruction residual is controlled by basis consistency; restatement of Theorem~\ref{thm:reconstruction_error}]
\label{app:thm_reconstruction_error}
The minimum basis alignment residual satisfies
\begin{equation}
\min_{T\in\mathcal O(r_i)}
\|Q_i^tT-Q_{\mathrm{ref},i}^{t-1}\|_F^2
=
2r_i(1-c_i^t),
\label{eq:app_consistency_residual}
\end{equation}
and the basis-consistent local matrix update satisfies
\begin{equation}
\|\Delta W_i^t-{\Delta \widetilde W}_i^t\|_F
\le
s\|\bar B_i^t\|_2
\sqrt{2r_i(1-c_i^t)}.
\label{eq:app_reconstruction_bound}
\end{equation}
\end{theorem}

\begin{proof}
We prove the two claims separately.

\paragraph{Step 1: exact alignment residual.}
For any $T\in\mathcal O(r_i)$,
\begin{align}
\|Q_i^tT-Q_{\mathrm{ref},i}^{t-1}\|_F^2
&=
\operatorname{tr}
\left[
(Q_i^tT-Q_{\mathrm{ref},i}^{t-1})^\top
(Q_i^tT-Q_{\mathrm{ref},i}^{t-1})
\right]
\\
&=
\operatorname{tr}\left(T^\top(Q_i^t)^\top Q_i^tT\right)
+
\operatorname{tr}\left((Q_{\mathrm{ref},i}^{t-1})^\top Q_{\mathrm{ref},i}^{t-1}\right)
\\
&\quad
-
2\operatorname{tr}\left(T^\top(Q_i^t)^\top Q_{\mathrm{ref},i}^{t-1}\right).
\end{align}
Since both $Q_i^t$ and $Q_{\mathrm{ref},i}^{t-1}$ have orthonormal columns,
$
(Q_i^t)^\top Q_i^t=I_{r_i}$ and
$(Q_{\mathrm{ref},i}^{t-1})^\top Q_{\mathrm{ref},i}^{t-1}=I_{r_i}$. Thus,
\begin{equation}
\|Q_i^tT-Q_{\mathrm{ref},i}^{t-1}\|_F^2
=
2r_i
-
2\operatorname{tr}(T^\top M_i^t),
\qquad
M_i^t:=(Q_i^t)^\top Q_{\mathrm{ref},i}^{t-1}.
\label{eq:app_proc_reduced}
\end{equation}
Let $M_i^t=U_i^t\Sigma_i^t(V_i^t)^\top$. By von Neumann's trace inequality,
\[
\operatorname{tr}(T^\top M_i^t)
\le
\operatorname{tr}(\Sigma_i^t)
\]
for every orthogonal $T$, with equality attained by $T_i^t=U_i^t(V_i^t)^\top$. Therefore,
\begin{align}
\min_{T\in\mathcal O(r_i)}
\|Q_i^tT-Q_{\mathrm{ref},i}^{t-1}\|_F^2
&=
2r_i-2\operatorname{tr}(\Sigma_i^t)
=
2r_i(1-c_i^t).
\end{align}

\paragraph{Step 2: matrix reconstruction residual.}
For any $T\in\mathcal O(r_i)$, define $\Delta W_i^t(T):=s\bar B_i^tT(Q_{\mathrm{ref},i}^{t-1})^\top$. Since $TT^\top=I_{r_i}$,
\begin{align}
\Delta W_i^t
&=
s\bar B_i^t(Q_i^t)^\top
=
s\bar B_i^tTT^\top(Q_i^t)^\top
=
s\bar B_i^tT(Q_i^tT)^\top.
\end{align}
Hence
\begin{align}
\Delta W_i^t-\Delta W_i^t(T)
&=
s\bar B_i^tT(Q_i^tT)^\top
-
s\bar B_i^tT(Q_{\mathrm{ref},i}^{t-1})^\top
\\
&=
s\bar B_i^tT
(Q_i^tT-Q_{\mathrm{ref},i}^{t-1})^\top.
\end{align}
Taking Frobenius norms and using $\|XY\|_F\le \|X\|_2\|Y\|_F$,
\begin{align}
\|\Delta W_i^t-\Delta W_i^t(T)\|_F
&\le
s\|\bar B_i^tT\|_2
\|Q_i^tT-Q_{\mathrm{ref},i}^{t-1}\|_F
\\
&=
s\|\bar B_i^t\|_2
\|Q_i^tT-Q_{\mathrm{ref},i}^{t-1}\|_F,
\end{align}
where the last equality uses the orthogonality of $T$. Setting $T=T_i^t$ and using Step 1,
\[
\|\Delta W_i^t-{\Delta \widetilde W}_i^t\|_F
\le
s\|\bar B_i^t\|_2
\sqrt{2r_i(1-c_i^t)}.
\]
\end{proof}

\subsection{Consistency-Aware Weighting (Assumption~\ref{assump:kappa} and Theorem~\ref{thm:weight_tightening})}
\label{app:proof_weight_tightening}

\begin{assumption}[Uniform coefficient-radius bound; restatement of Assumption~\ref{assump:kappa}]
\label{app:assump_kappa}
There exists $\kappa_t>0$ such that
\begin{equation}
s\|\bar B_i^t\|_2\sqrt{r_i}
\le
\kappa_t
\label{eq:app_kappa_bound}
\end{equation}
for every $i\in\mathcal S^t$.
\end{assumption}

\begin{theorem}[Consistency-aware weighting tightens a reconstruction bound; restatement of Theorem~\ref{thm:weight_tightening}]
\label{app:thm_weight_tightening}
Let
\begin{equation}
b_i^t
:=
\kappa_t\sqrt{2(1-c_i^t)}.
\label{eq:app_bi_def}
\end{equation}
Under Assumption~\ref{app:assump_kappa},
\begin{equation}
\|\Delta W_i^t-{\Delta \widetilde W}_i^t\|_F
\le
b_i^t
\label{eq:app_per_client_bi}
\end{equation}
for every $i\in\mathcal S^t$. If $\sum_{j\in\mathcal S^t}p_j^t\,c_j^t>0$, then
\begin{equation}
\sum_{i\in\mathcal S^t}
\alpha_i^t\,b_i^t
\le
\sum_{i\in\mathcal S^t}
p_i^t\,b_i^t.
\label{eq:app_weight_tightening}
\end{equation}
\end{theorem}

\begin{proof}
By Theorem~\ref{app:thm_reconstruction_error},
\[
\|\Delta W_i^t-{\Delta \widetilde W}_i^t\|_F
\le
s\|\bar B_i^t\|_2
\sqrt{2r_i(1-c_i^t)}.
\]
Using Assumption~\ref{app:assump_kappa},
\begin{align}
s\|\bar B_i^t\|_2
\sqrt{2r_i(1-c_i^t)}
&=
\sqrt{2(1-c_i^t)}
\cdot
s\|\bar B_i^t\|_2\sqrt{r_i}
\le
\kappa_t\sqrt{2(1-c_i^t)}
=
b_i^t.
\end{align}
This proves Eq.~\eqref{eq:app_per_client_bi}.

It remains to prove the weighted inequality. Define $b(c)=\kappa_t\sqrt{2(1-c)}$, which is non-increasing on $[0,1]$. Let $I$ and $J$ be two independent random indices drawn from the distribution $\{p_i\}_{i\in\mathcal S^t}$. Since $c\mapsto c$ is non-decreasing and $c\mapsto b(c)$ is non-increasing,
\[
(c_I^t-c_J^t)(b_I^t-b_J^t)\le 0.
\]
Taking expectation,
\begin{equation}
\mathbb E[(c_I^t-c_J^t)(b_I^t-b_J^t)]\le 0.
\label{eq:app_negative_corr}
\end{equation}
Expanding and using independence of $I,J$,
\begin{align}
\mathbb E[(c_I^t-c_J^t)(b_I^t-b_J^t)]
&=
2\mathbb E[c_I^tb_I^t]
-
2\mathbb E[c_I^t]\mathbb E[b_I^t].
\end{align}
Thus, $\mathbb E[c_I^tb_I^t]\le\mathbb E[c_I^t]\mathbb E[b_I^t]$, i.e.,
\begin{equation}
\sum_{i\in\mathcal S^t}
p_i\,c_i^t\,b_i^t
\le
\left(
\sum_{i\in\mathcal S^t}p_i^t\,c_i^t
\right)
\left(
\sum_{i\in\mathcal S^t}p_i^t\,b_i^t
\right).
\label{eq:app_cov_ineq}
\end{equation}
Since $\sum_i p_i^t\,c_i^t>0$, dividing by this positive denominator gives
\[
\frac{
\sum_i p_i^t\,c_i^t\,b_i^t
}{
\sum_i p_i^t\,c_i^t
}
\le
\sum_i p_i^t\,b_i^t,
\]
i.e., $\sum_i\alpha_i^t b_i^t\le\sum_i p_i^t\,b_i^t$.
\end{proof}

\paragraph{Remark on equality.}
If all consistency scores are equal on the support of $p$, then $\alpha_i^t=p_i^t$ and Eq.~\eqref{eq:app_weight_tightening} holds with equality. If all $p_i^t>0$ and the consistency scores are not all equal, then the inequality is strict.

\subsection{Variational Characterization of the Weighting Rule (Remark~\ref{rem:variational})}
\label{app:variational_weighting}

This subsection formalizes Remark~\ref{rem:variational} from the main text, showing that the consistency-aware weighting rule $\alpha_i^t$ admits a variational characterization as a KL-regularized log-consistency maximizer.

\begin{proposition}[Variational characterization of $\alpha_i^t$]
\label{app:prop_variational_weighting}
Assume $p_i^t>0$ and $c_i^t>0$ for every $i\in\mathcal S^t$. Then
\begin{equation}
\alpha_i^t
=
\frac{p_i^t\,c_i^t}{\sum_{j\in\mathcal S^t}p_j^t\,c_j^t}
\label{eq:app_variational_rule}
\end{equation}
is the unique maximizer of
\begin{equation}
\max_{\alpha\in\Delta_{|\mathcal S^t|}}
\left\{
\sum_{i\in\mathcal S^t}\alpha_i\log c_i^t
-
\mathrm{KL}(\alpha\|p)
\right\},
\label{eq:app_variational_obj}
\end{equation}
where $\mathrm{KL}(\alpha\|p)=\sum_i\alpha_i\log\frac{\alpha_i}{p_i}$.
\end{proposition}

\begin{proof}
The Lagrangian is
\[
\mathcal L(\alpha,\lambda)
=
\sum_i\alpha_i\log c_i^t
-
\sum_i\alpha_i\log\frac{\alpha_i}{p_i}
+
\lambda\left(\sum_i\alpha_i-1\right).
\]
Differentiating with respect to $\alpha_i$,
\[
\frac{\partial\mathcal L}{\partial\alpha_i}
=
\log c_i^t
-
\left(
\log\frac{\alpha_i}{p_i}+1
\right)
+
\lambda.
\]
Setting the derivative to zero, $\log\frac{\alpha_i}{p_i}=\log c_i^t+\lambda-1$, so $\alpha_i=e^{\lambda-1}p_i^t\,c_i^t$. The simplex constraint gives $e^{\lambda-1}=1/\sum_jp_j\,c_j^t$, hence
\[
\alpha_i
=
\frac{p_i^t\,c_i^t}{\sum_jp_j^t\,c_j^t}.
\]
The objective in Eq.~\eqref{eq:app_variational_obj} is a linear function of $\alpha$ minus $\mathrm{KL}(\alpha\|p)$. Since $\mathrm{KL}(\alpha\|p)$ is strictly convex on the relative interior of the simplex when $p_i^t>0$, the objective is strictly concave, and the stationary point is the unique maximizer.
\end{proof}

\subsection{Idealized Optimization Interpretation (Assumption~\ref{assump:conv} and Theorem~\ref{thm:convergence})}
\label{app:proof_convergence}

\begin{assumption}[Idealized smooth stochastic descent model; restatement of Assumption~\ref{assump:conv}]
\label{app:assump_conv}
The objective $F$ is $L$-smooth, and the raw aggregate $G^t=\sum_{i\in\mathcal S^t}\alpha_i^t\Delta W_i^t$ satisfies
\begin{equation}
\mathbb E[G^t\mid W^t]
=
\nabla F(W^t),
\qquad
\mathbb E[
\|G^t-\nabla F(W^t)\|_F^2
\mid W^t
]
\le
\sigma^2.
\label{eq:app_conv_assump}
\end{equation}
\end{assumption}

\begin{theorem}[Convergence with reconstruction bias; restatement of Theorem~\ref{thm:convergence}]
\label{app:thm_convergence}
Define
\begin{equation}
\widetilde G^t
=
\sum_{i\in\mathcal S^t}
\alpha_i^t{\Delta \widetilde W}_i^t,
\qquad
e^t
=
\widetilde G^t-G^t.
\label{eq:app_G_defs}
\end{equation}
Consider the idealized server update
\begin{equation}
W^{t+1}
=
W^t-\eta\widetilde G^t
\label{eq:app_server_update}
\end{equation}
with $\eta\le 1/(4L)$. Under Assumption~\ref{app:assump_conv},
\begin{equation}
\frac{1}{T}
\sum_{t=0}^{T-1}
\mathbb E\|\nabla F(W^t)\|_F^2
\le
\frac{4(F(W^0)-F^\star)}{\eta T}
+
4L\eta\sigma^2
+
\frac{2+4L\eta}{T}
\sum_{t=0}^{T-1}
\mathbb E\|e^t\|_F^2.
\label{eq:app_convergence}
\end{equation}
Moreover,
\begin{equation}
\|e^t\|_F
\le
\sum_{i\in\mathcal S^t}
\alpha_i^t\,
s\|\bar B_i^t\|_2
\sqrt{2r_i(1-c_i^t)}.
\label{eq:app_bias_control}
\end{equation}
\end{theorem}

\begin{proof}
We first bound the reconstruction perturbation. By definition,
\begin{align}
e^t
&=
\sum_{i\in\mathcal S^t}
\alpha_i^t
(
{\Delta \widetilde W}_i^t-\Delta W_i^t
).
\end{align}
Using the triangle inequality and $\alpha_i^t\ge 0$,
\[
\|e^t\|_F
\le
\sum_{i\in\mathcal S^t}
\alpha_i^t
\|{\Delta \widetilde W}_i^t-\Delta W_i^t\|_F.
\]
By Theorem~\ref{app:thm_reconstruction_error},
\[
\|{\Delta \widetilde W}_i^t-\Delta W_i^t\|_F
\le
s\|\bar B_i^t\|_2
\sqrt{2r_i(1-c_i^t)},
\]
which proves Eq.~\eqref{eq:app_bias_control}.

We now prove the convergence bound. Since $F$ is $L$-smooth,
\begin{align}
F(W^{t+1})
&\le
F(W^t)
+
\langle\nabla F(W^t),W^{t+1}-W^t\rangle
+
\frac{L}{2}\|W^{t+1}-W^t\|_F^2
\\
&=
F(W^t)
-
\eta\langle\nabla F(W^t),\widetilde G^t\rangle
+
\frac{L\eta^2}{2}\|\widetilde G^t\|_F^2.
\end{align}
Using $\widetilde G^t=G^t+e^t$,
\begin{align}
F(W^{t+1})
\le\;&
F(W^t)
-
\eta\langle\nabla F(W^t),G^t\rangle
-
\eta\langle\nabla F(W^t),e^t\rangle
\\
&+
\frac{L\eta^2}{2}
\|G^t+e^t\|_F^2.
\end{align}
Using $\|G^t+e^t\|_F^2\le 2\|G^t\|_F^2+2\|e^t\|_F^2$,
\begin{align}
F(W^{t+1})
\le\;&
F(W^t)
-
\eta\langle\nabla F(W^t),G^t\rangle
-
\eta\langle\nabla F(W^t),e^t\rangle
\\
&+
L\eta^2\|G^t\|_F^2
+
L\eta^2\|e^t\|_F^2.
\end{align}
Taking conditional expectation given $W^t$ and using Assumption~\ref{app:assump_conv},
\[
\mathbb E[
\langle\nabla F(W^t),G^t\rangle
\mid W^t
]
=
\|\nabla F(W^t)\|_F^2,
\]
and
\begin{align}
\mathbb E[\|G^t\|_F^2\mid W^t]
&=
\|\nabla F(W^t)\|_F^2
+
\mathbb E[
\|G^t-\nabla F(W^t)\|_F^2
\mid W^t
]
\\
&\le
\|\nabla F(W^t)\|_F^2+\sigma^2.
\end{align}
Thus,
\begin{align}
\mathbb E[F(W^{t+1})\mid W^t]
\le\;&
F(W^t)
-
\eta\|\nabla F(W^t)\|_F^2
-
\eta
\mathbb E[
\langle\nabla F(W^t),e^t\rangle
\mid W^t
]
\\
&+
L\eta^2\|\nabla F(W^t)\|_F^2
+
L\eta^2\sigma^2
+
L\eta^2
\mathbb E[\|e^t\|_F^2\mid W^t].
\end{align}
By Young's inequality,
\[
-\langle\nabla F(W^t),e^t\rangle
\le
\frac{1}{2}\|\nabla F(W^t)\|_F^2
+
\frac{1}{2}\|e^t\|_F^2.
\]
Substituting,
\begin{align}
\mathbb E[F(W^{t+1})\mid W^t]
\le\;&
F(W^t)
-
\eta\left(\frac12-L\eta\right)
\|\nabla F(W^t)\|_F^2
\\
&+
L\eta^2\sigma^2
+
\left(\frac{\eta}{2}+L\eta^2\right)
\mathbb E[\|e^t\|_F^2\mid W^t].
\end{align}
If $\eta\le 1/(4L)$, then $\frac12-L\eta\ge \frac14$, so
\[
\frac{\eta}{4}
\|\nabla F(W^t)\|_F^2
\le
F(W^t)
-
\mathbb E[F(W^{t+1})\mid W^t]
+
L\eta^2\sigma^2
+
\left(\frac{\eta}{2}+L\eta^2\right)
\mathbb E[\|e^t\|_F^2\mid W^t].
\]
Taking full expectation and summing over $t=0,\ldots,T-1$,
\begin{align}
\frac{\eta}{4}
\sum_{t=0}^{T-1}
\mathbb E\|\nabla F(W^t)\|_F^2
\le\;&
F(W^0)-\mathbb E[F(W^T)]
+
TL\eta^2\sigma^2
\\
&+
\left(\frac{\eta}{2}+L\eta^2\right)
\sum_{t=0}^{T-1}
\mathbb E\|e^t\|_F^2.
\end{align}
Using $F(W^T)\ge F^\star$ and dividing by $\eta T/4$,
\[
\frac{1}{T}
\sum_{t=0}^{T-1}
\mathbb E\|\nabla F(W^t)\|_F^2
\le
\frac{4(F(W^0)-F^\star)}{\eta T}
+
4L\eta\sigma^2
+
\frac{2+4L\eta}{T}
\sum_{t=0}^{T-1}
\mathbb E\|e^t\|_F^2.
\]
\end{proof}

\textbf{Why there is no projection bias.}
By Proposition~\ref{app:prop_shared_coord}, $\widetilde G^t=sC_g^t(Q_g^{t-1})^\top$ for some $C_g^t\in\mathbb R^{m\times r_{\max}}$, so $\operatorname{rank}(\widetilde G^t)\le r_{\max}$. Combined with Propositions~\ref{app:prop_exact_svd_refresh} and~\ref{app:prop_gauge_invariance}, the rank-$r_{\max}$ SVD refresh and gauge fixing exactly represent $\widetilde G^t$. They change only the factor representation and the next-round global reference basis; they do not add a separate low-rank projection error.

\section{Additional Experimental Results}
\label{app:cr_experiments}

\subsection{Results across Independent Runs}
\label{app:cr_reproducibility}

We compare FedAbA and Fed-PLoRA over three independent matched
runs under the settings of the main evaluation.
The repeated-run study covers MRPC F1, RTE accuracy, MMLU after
Dolly-15K tuning, MT-Bench after Alpaca tuning, and Natural
Instructions ROUGE-L and ROUGE-1.
For metrics evaluated at every communication round, the final
score of each run is averaged over the last five rounds.
Table~\ref{tab:cr_repeated_runs} reports the mean and sample
standard deviation across the three run-level scores.
Thus, the reported standard deviation measures variation across
independent runs, rather than variation across the five
evaluation rounds within a run.

\begin{table}[hpt!]
    \centering
    \small
    \setlength{\tabcolsep}{5pt}
    \caption{
        Repeated-run comparison: mean $\pm$ sample standard
        deviation over three independent matched runs.
        Dolly-15K and Alpaca denote training datasets, evaluated
        with MMLU and MT-Bench, respectively.
    }
    \label{tab:cr_repeated_runs}
    \resizebox{0.95\linewidth}{!}{%
    \begin{tabular}{lrrrrrr}
        \toprule
        Method
        & \shortstack{MRPC\\F1}
        & \shortstack{RTE\\Acc.}
        & \shortstack{Dolly-15K\\MMLU}
        & \shortstack{Alpaca\\MT-Bench}
        & \shortstack{NI\\ROUGE-L}
        & \shortstack{NI\\ROUGE-1} \\
        \midrule
        Fed-PLoRA
        & $89.42\pm0.45$
        & $81.63\pm1.10$
        & $47.17\pm0.58$
        & $4.21\pm0.12$
        & $57.98\pm0.46$
        & $65.66\pm0.48$ \\
        FedAbA
        & $\mathbf{90.28}\pm0.64$
        & $\mathbf{83.28}\pm0.82$
        & $\mathbf{47.87}\pm0.40$
        & $\mathbf{4.31}\pm0.14$
        & $\mathbf{58.37}\pm0.30$
        & $\mathbf{66.05}\pm0.63$ \\
        \bottomrule
    \end{tabular}%
    }
\end{table}

\subsection{Performance on Low-Consistency Clients}
\label{app:cr_low_consistency}

We examine performance on clients whose uploaded bases are less
aligned with the reference in the Natural Instructions
task-as-client setting. We rank clients by their average
consistency in the default FedAbA run and select the lowest
20\%. The same fixed subset is used to evaluate both FedAbA
and Fed-PLoRA.
Table~\ref{tab:cr_low_consistency} reports ROUGE-L and ROUGE-1
over all clients and over this subset.

\begin{table}[hpt!]
    \centering
    \small
    \setlength{\tabcolsep}{5pt}
    \caption{
        Natural Instructions results over all clients and the
        fixed lowest-consistency 20\% subset.
    }
    \label{tab:cr_low_consistency}
    \resizebox{0.6\linewidth}{!}{%
    \begin{tabular}{lrrrr}
        \toprule
        Method
        & \multicolumn{2}{c}{All clients}
        & \multicolumn{2}{c}{Lowest-consistency 20\%} \\
        \cmidrule(lr){2-3}
        \cmidrule(lr){4-5}
        & ROUGE-L
        & ROUGE-1
        & ROUGE-L
        & ROUGE-1 \\
        \midrule
        Fed-PLoRA
        & 58.1
        & 65.8
        & 54.8
        & 62.4 \\
        FedAbA
        & \textbf{58.4}
        & \textbf{66.1}
        & \textbf{55.1}
        & \textbf{62.6} \\
        \bottomrule
    \end{tabular}%
    }
\end{table}

Compared with Fed-PLoRA, FedAbA improves all-client ROUGE-L and
ROUGE-1 by $0.3$ points each.
On the fixed low-consistency subset, the corresponding
improvements are $0.3$ and $0.2$ points.
The improvement over Fed-PLoRA therefore also appears in the
average performance of this subset.
This comparison evaluates the complete methods and does not
isolate the effect of residual removal or establish improvement
for every individual client.

\subsection{Additional Backbone Evaluation}
\label{app:cr_backbones}

We evaluate additional backbones while retaining the Dolly-15K
high-rank-heterogeneity protocol and the MMLU evaluation used in
the main experiment.
Qwen3-4B tests a model from another architecture family, while
LLaMA-2-13B tests a larger model in the LLaMA-2 family.
Table~\ref{tab:cr_backbones} includes the original LLaMA-2-7B
results for reference and compares FedAbA with FlexLoRA,
ECLoRA, Fed-PLoRA, and a FedAbA variant without basis-consistent
reconstruction.

\begin{table}[hpt!]
    \centering
    \small
    \setlength{\tabcolsep}{5pt}
    \caption{
        Backbone comparison after federated fine-tuning on
        Dolly-15K, evaluated with MMLU.
    }
    \label{tab:cr_backbones}
    \resizebox{0.9\linewidth}{!}{%
    \begin{tabular}{lrrr}
        \toprule
        Method
        & LLaMA-2-7B
        & Qwen3-4B
        & LLaMA-2-13B \\
        \midrule
        FlexLoRA
        & 45.9
        & 57.0
        & 51.7 \\
        ECLoRA
        & 46.1
        & 57.2
        & 51.9 \\
        Fed-PLoRA
        & 47.2
        & 58.3
        & 53.0 \\
        FedAbA w/o basis-consistent reconstruction
        & 46.9
        & 58.0
        & 52.7 \\
        FedAbA
        & \textbf{47.8}
        & \textbf{59.0}
        & \textbf{53.7} \\
        \bottomrule
    \end{tabular}%
    }
\end{table}

\section{Broader Impact}
\label{sec:broader-impact}
FedAbA aims to improve federated LLM fine-tuning under heterogeneous client capacities by stabilizing rank-prefix dispatch through basis-consistent aggregation, which may benefit privacy-preserving adaptation in domains where data ownership or regulation limits data sharing, particularly for resource-constrained clients that can only afford low-rank adapters. 

\section{Limitations}
\label{sec:limitations}
Despite its effectiveness, FedAbA has several limitations.
First, the default aggregation operates within the span of the warm-started global basis: the SVD refresh and gauge fixing evolve the global basis only by an orthogonal transform of the warm-start basis, which stabilizes rank-prefix semantics across rounds but cannot introduce genuinely new basis directions if the warm-start span is unrepresentative.
Second, FedAbA aggregates only the basis-consistent reconstruction of each client's local matrix update and discards the corresponding residual; when the consistency score is small for low-rank clients in early rounds, this trades a non-negligible residual signal for cross-round prefix stability, and a principled mechanism for selectively recovering it without reintroducing inconsistency is left to future work.


\newpage
\section*{NeurIPS Paper Checklist}

\begin{enumerate}

\item {\bf Claims}
    \item[] Question: Do the main claims made in the abstract and introduction accurately reflect the paper's contributions and scope?
    \item[] Answer: \answerYes{} 
    \item[] Justification: The main claims made in the abstract and introduction accurately reflect the paper’s contributions and scope.

\item {\bf Limitations}
    \item[] Question: Does the paper discuss the limitations of the work performed by the authors?
    \item[] Answer: \answerYes{} 
    \item[] Justification: We detail the limitations of the framework in the appendix.

\item {\bf Theory assumptions and proofs}
    \item[] Question: For each theoretical result, does the paper provide the full set of assumptions and a complete (and correct) proof?
    \item[] Answer: \answerYes{}
    \item[] Justification: The complete assumptions, lemmas, and proofs are provided in the appendix.

    \item {\bf Experimental result reproducibility}
    \item[] Question: Does the paper fully disclose all the information needed to reproduce the main experimental results of the paper to the extent that it affects the main claims and/or conclusions of the paper (regardless of whether the code and data are provided or not)?
    \item[] Answer: \answerYes{} 
    \item[] Justification: The paper provides comprehensive information necessary to reproduce the main experimental results, ensuring transparency and replicability of the findings.

\item {\bf Open access to data and code}
    \item[] Question: Does the paper provide open access to the data and code, with sufficient instructions to faithfully reproduce the main experimental results, as described in supplemental material?
    \item[] Answer: \answerYes{} 
    \item[] Justification: Our code is available at https://anonymous.4open.science/r/FedAbA-6774

\item {\bf Experimental setting/details}
    \item[] Question: Does the paper specify all the training and test details (e.g., data splits, hyperparameters, how they were chosen, type of optimizer) necessary to understand the results?
    \item[] Answer: \answerYes{} 
    \item[] Justification: We specify all the training and test details in the appendix.

\item {\bf Experiment statistical significance}
    \item[] Question: Does the paper report error bars suitably and correctly defined or other appropriate information about the statistical significance of the experiments?
    \item[] Answer: \answerYes{} 
    \item[] Justification: The results are reproducible with low errors.

\item {\bf Experiments compute resources}
    \item[] Question: For each experiment, does the paper provide sufficient information on the computer resources (type of compute workers, memory, time of execution) needed to reproduce the experiments?
    \item[] Answer: \answerYes{} 
    \item[] Justification: We provide the information on the computer resources in appendix.
    
\item {\bf Code of ethics}
    \item[] Question: Does the research conducted in the paper conform, in every respect, with the NeurIPS Code of Ethics \url{https://neurips.cc/public/EthicsGuidelines}?
    \item[] Answer: \answerYes{} 
    \item[] Justification: The research conducted in this paper strictly adheres to the NeurIPS Code of Ethics. We ensured that all data used in our study was obtained with proper consent, and we adhered to principles of fairness, transparency, and accountability. The research does not involve any form of discrimination or harm to individuals or groups, and we made sure to conduct the study with full respect for privacy and confidentiality.

\item {\bf Broader impacts}
    \item[] Question: Does the paper discuss both potential positive societal impacts and negative societal impacts of the work performed?
    \item[] Answer: \answerYes{} 
    \item[] Justification: The paper discusses the potential positive societal impacts of the work performed, highlighting the benefits and advancements it can bring to the field and society at large in the appendix.
    
\item {\bf Safeguards}
    \item[] Question: Does the paper describe safeguards that have been put in place for responsible release of data or models that have a high risk for misuse (e.g., pre-trained language models, image generators, or scraped datasets)?
    \item[] Answer: \answerNA{} 
    \item[] Justification: This paper presents no such risks.

\item {\bf Licenses for existing assets}
    \item[] Question: Are the creators or original owners of assets (e.g., code, data, models), used in the paper, properly credited and are the license and terms of use explicitly mentioned and properly respected?
    \item[] Answer: \answerYes{} 
    \item[] Justification: We have appropriately credited and cited the creators or original owners of all assets, including code, data, and papers, used in our work. Additionally, the licenses and terms of use for these assets have been explicitly stated and fully respected.

\item {\bf New assets}
    \item[] Question: Are new assets introduced in the paper well documented and is the documentation provided alongside the assets?
    \item[] Answer: \answerNA{}  
    \item[] Justification: The paper does not introduce any new assets.

\item {\bf Crowdsourcing and research with human subjects}
    \item[] Question: For crowdsourcing experiments and research with human subjects, does the paper include the full text of instructions given to participants and screenshots, if applicable, as well as details about compensation (if any)? 
    \item[] Answer: \answerNA{} 
    \item[] Justification: This paper does not involve crowdsourcing or research involving human subjects.

\item {\bf Institutional review board (IRB) approvals or equivalent for research with human subjects}
    \item[] Question: Does the paper describe potential risks incurred by study participants, whether such risks were disclosed to the subjects, and whether Institutional Review Board (IRB) approvals (or an equivalent approval/review based on the requirements of your country or institution) were obtained?
    \item[] Answer: \answerNA{} 
    \item[] Justification: This paper does not involve crowdsourcing or research with human subjects.

\item {\bf Declaration of LLM usage}
    \item[] Question: Does the paper describe the usage of LLMs if it is an important, original, or non-standard component of the core methods in this research? Note that if the LLM is used only for writing, editing, or formatting purposes and does \emph{not} impact the core methodology, scientific rigor, or originality of the research, declaration is not required.
    \item[] Answer: \answerNA{} 
    \item[] Justification:  The core method development in this research does not involve LLMs as any important, original, or non-standard components.

\end{enumerate}

\end{document}